\documentclass[letterpaper]{article} 
\usepackage{aaai25}  
\usepackage{times}  
\usepackage{helvet}  
\usepackage{courier}  
\usepackage[hyphens]{url}  
\usepackage{graphicx} 
\usepackage{natbib}  
\usepackage{caption} 
\usepackage{algorithm}
\usepackage{algorithmic}
\usepackage{multirow}
\usepackage{amsmath}
\usepackage{amssymb}
\usepackage[table]{xcolor}

\usepackage{newfloat}
\usepackage{listings}
\DeclareCaptionStyle{ruled}{labelfont=normalfont,labelsep=colon,strut=off} 
\floatstyle{ruled}
\newfloat{listing}{tb}{lst}{}
\floatname{listing}{Listing}
\title{DVP-MVS: Synergize Depth-Edge and Visibility Prior for Multi-View Stereo}

\author{
    Zhenlong Yuan\textsuperscript{\rm 1}, 
    Jinguo Luo\textsuperscript{\rm 2}, 
    Fei Shen\textsuperscript{\rm 3}, 
    Zhaoxin Li\textsuperscript{\rm 4, \rm 5}, 
    Cong Liu\textsuperscript{\rm 2}, 
    Tianlu Mao\textsuperscript{\rm 1}\thanks{Corresponding Author.}, 
    Zhaoqi Wang\textsuperscript{\rm 1}
}
\affiliations{
    \textsuperscript{\rm 1}Institute of Computing Technology, Chinese Academy of Sciences

    \textsuperscript{\rm 2}Harbin Institute of Technology, Shenzhen

    \textsuperscript{\rm 3}Nanjing University of Science and Technology, Nanjing
    
    \textsuperscript{\rm 4}Agricultural Information Institute, Chinese Academy of Agricultural Sciences
    
    \textsuperscript{\rm 5}Key Laboratory of Agricultural Big Data, Ministry of Agriculture and Rural Affairs

    yuanzhenlong21b@ict.ac.cn, 23s153135@stu.hit.edu.cn, feishen@njust.edu.cn, 

    cszli@hotmail.com, liucong@stu.hit.edu.cn, \{ltm, zqwang\}@ict.ac.cn
}

\usepackage{bibentry}

\begin{document}

\maketitle

\begin{abstract} 
Patch deformation-based methods have recently exhibited substantial effectiveness in multi-view stereo, due to the incorporation of deformable and expandable perception to reconstruct textureless areas. 
However, such approaches typically focus on exploring correlative reliable pixels to alleviate match ambiguity during patch deformation, but ignore the deformation instability caused by mistaken edge-skipping and visibility occlusion, leading to potential estimation deviation. 
To remedy the above issues, we propose DVP-MVS, which innovatively synergizes depth-edge aligned and cross-view prior for robust and visibility-aware patch deformation.
Specifically, to avoid unexpected edge-skipping, we first utilize Depth Anything V2 followed by the Roberts operator to initialize coarse depth and edge maps respectively, both of which are further aligned through an erosion-dilation strategy to generate fine-grained homogeneous boundaries for guiding patch deformation.
In addition, we reform view selection weights as visibility maps and restore visible areas by cross-view depth reprojection, then regard them as cross-view prior to facilitate visibility-aware patch deformation. 
Finally, we improve propagation and refinement with multi-view geometry consistency by introducing aggregated visible hemispherical normals based on view selection and local projection depth differences based on epipolar lines, respectively.
Extensive evaluations on ETH3D and Tanks \& Temples benchmarks demonstrate that our method can achieve state-of-the-art performance with excellent robustness and generalization.

\end{abstract}

\section{Introduction}
Multi-View Stereo (MVS), as a core task in computer vision, aims to dense the geometry representation of the scene or object through certain overlapping photographs from different viewpoints.
It has been employed in various fields including scene reconstruction \cite{YZF_2, YZF_1}, image denoising \cite{Yang2021, Yang2024c, Li2022b}, pose estimation \cite{Shen2024, Shen2024b, Shen2024c, Shen2024a}, etc.
In recent years, the emergence of plentiful imaginative ideas \cite{Gan2021, Gan2024a, GUAN1, GUAN2} tremendously boosts its performance among various benchmarks \cite{ETH3D, Blendedmvs, XIMAGENET}. These innovations can be roughly categorized into learning-based MVS and traditional MVS. 

\begin{figure}
    \centering \includegraphics[width=\linewidth]{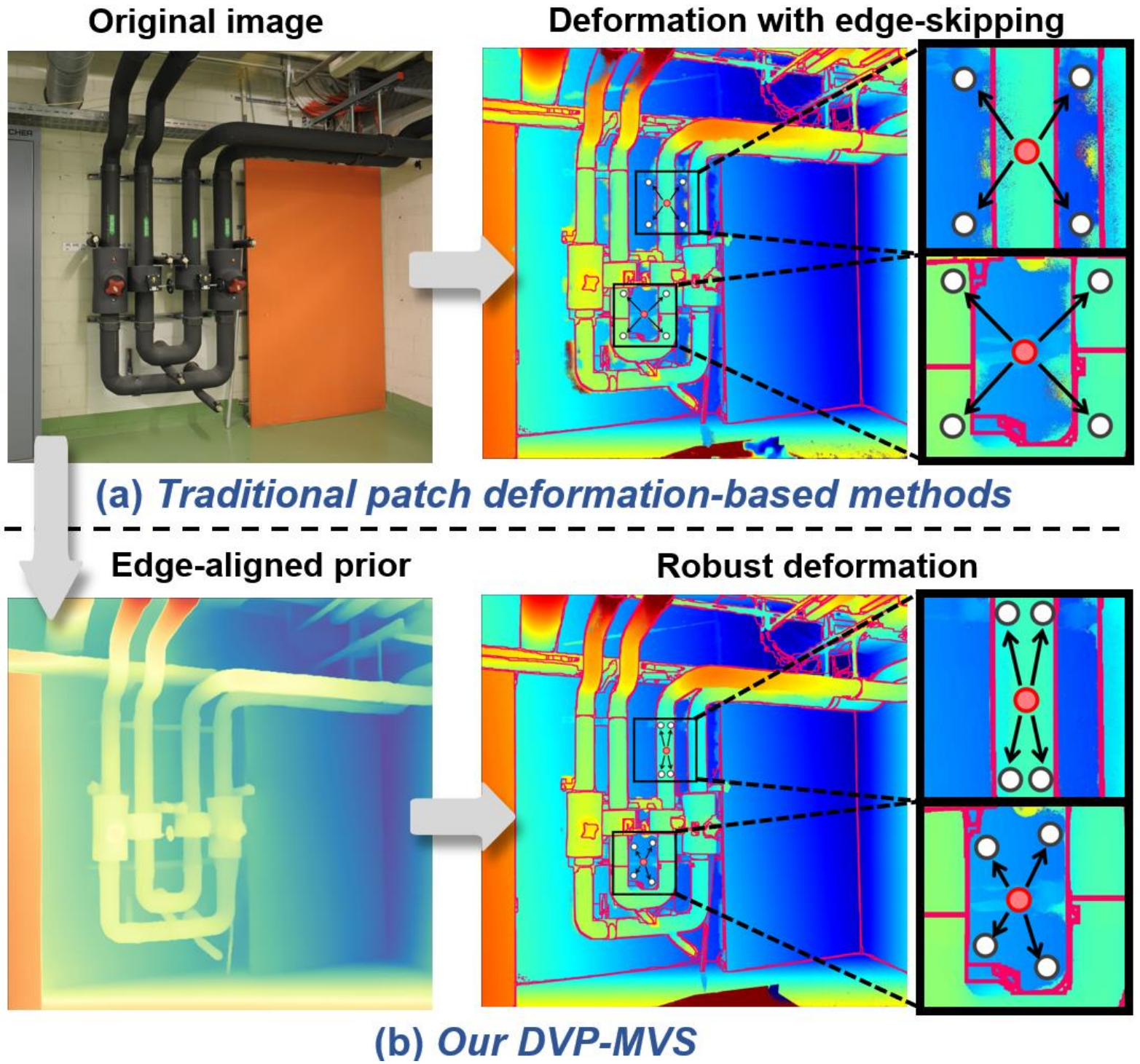}
    \caption{
    Comparison between patch deformation-based methods and ours. Edge-skipping causes other methods (a) incorrectly select reliable but depth-discontinuous gray pixels for patch deformation of the central unreliable red pixel. While our DVP-MVS (b) leverages depth-edge aligned prior to guarantee deformed patches within homogeneous areas.
    }
    \label{fig1}
\end{figure}

Learning-based MVS leverage convolutional neural networks to extract high-dimensional features for reconstruction, while they demand large training datasets and show inferior generalization abilities.
On the contrary, traditional MVS are extended from the PatchMatch algorithm \cite{PM}, which propagates appropriate hypotheses from neighboring pixels and generates refined hypotheses to construct solution space, then elect the optimal hypotheses through a criterion defined by the multi-view matching cost.
However, when the patch is located in textureless areas, its matching cost becomes untrustworthy because of lacking distinguishable features within the receptive field. 


To reconstruct textureless areas, two primary types of traditional MVS have been proposed: planarization-based and patch deformation-based methods. 
Planarization-based methods connect reliable pixels in well-textured areas to form regions that divide textureless areas, and then planarize these regions for reconstruction. 
Various strategies like superpixel \cite{TSAR-MVS}, triangulation \cite{ACMMP} and KD-tree \cite{HPM-MVS} are employed, but these methods typically suffer from the limited size of connected regions and are prone to deviations during planarization.

Differently, patch deformation-based methods perform deformation on fixed patches to introduce enough features for matching costs. 
E.g., API-MVS \cite{API-MVS} introduces entropy to increase the patch size and sampling intervals, while APD-MVS \cite{APD-MVS} adaptively searches for correlative reliable pixels around each unreliable pixel, then constructs multiple sub-patches centered on them for matching costs.
However, such methods primarily concentrate on exploring advanced reliable pixel searching strategies to alleviate match ambiguity, but ignore a critical constraint for deformation stability (i.e., depth-edge aligned prior).
As illustrated in Fig.~\ref{fig1}(a), when reconstructing the \textbf{unreliable} red pixels (i.e., pixels with matching ambiguity), traditional patch deformation-based methods typically segment their surrounding areas into multiple fixed-angle sectors, and then search a reliable gray pixel within each sector to form patches.
Nevertheless, during the process of searching for reliable pixels, unexpected edge-skipping caused by shadows or occlusions breaks the depth continuity principle, which may result in potential matching distortions.

Addressing this, we propose DVP-MVS, which innovatively synergizes depth-edge aligned and cross-view prior to facilitate robust and
visibility-aware patch deformation.
Specifically, we first utilize Depth Anything V2 \cite{depany2} followed by the Roberts operator to initialize coarse depth and edge maps, respectively. 
The former maps provide monocular depth information for \textbf{homogeneous areas} (i.e., areas with depth continuity) in the global level, but lack detailed edge information. 
In contrast, the latter maps offer information about rough edges and dispersed regions. 
Consequently, we introduce an erosion-dilation strategy to align both types of maps for acquiring fine-grained \textbf{homogeneous boundaries} (i.e., bounds for homogeneous areas) to guide patch deformation, guaranteeing deformed patches within depth-continuous areas as shown in Fig.~\ref{fig1}(b).

Furthermore, visibility occlusion is another crucial issue leading to unstable patch deformation.
To mitigate this, we initially construct visibility maps by reforming view selection weights in ACMMP~\cite{ACMMP}. 
Then, a post-verification algorithm based on coss-view depth reprojection is proposed for restoration of original-visible areas in visibility maps.
Considering the restored visibility maps as the cross-view prior, we iteratively update deformed patches, thereby achieving visibility-aware patch deformation.


Finally, we improve the propagation and refinement process by integrating multi-view geometric consistency. Instead of randomly generating hypotheses as in previous methods, we aggregate visible hemispherical normals to constrain the normal ranges and adopt local projection depth differences on epipolar lines to limit the depth intervals.
Evaluations on ETH3D and Tanks \& Temples benchmarks demonstrate that our method can achieve state-of-the-art performance with excellent robustness and generalization.

In summary, our contributions are four-fold:
\begin{itemize}
    \item For patch deformation-based MVS, we introduce depth-edge prior, which aligns coarse depth and edge information through an erosion-dilation strategy to generate fine-grained homogeneous boundaries for stable deformation.
    \item We construct visibility maps by reforming view selection weights and restoring visible areas with reprojection post-verification, which are then regarded as cross-view prior to facilitate visibility-aware patch deformation.
    \item Considering multi-view geometric consistency, we further improve the propagation and refinement process by introducing aggregated visible hemispherical normals and local projection depth differences on epipolar lines.
\end{itemize}

\section{Related Work}
\subsection{Traditional MVS Methods}
\subsubsection{PatchMatch MVS. } 
PatchMatch \cite{PM} aims to search for the approximate patch pairs between two images. 
PatchMatchStereo \cite{PMS} extends this idea to MVS, establishing the foundation for many subsequent methods. 
Gipuma \cite{Galliani} introduces a red-black checkerboard propagation to realize the parallelism of GPUs for acceleration. 
ACMM \cite{ACMM} further proposes the adaptive checkerboard sampling and joint view selection to reconstruct textureless areas. 
To further reconstruct textureless areas, ACMMP \cite{ACMMP} and HPM-MVS \cite{HPM-MVS} respectively leverage triangulation and KD-tree to extract planes for planarization.
Moreover, Pyramid \cite{Pyramid} and MG-MVS \cite{MG-MVS} introduce hierarchical architecture to indirectly enhance the receptive field for reconstruction. 
MAR-MVS \cite{MAR-MVS} introduces epipolar lines to analyze the geometry surface for optimal pixel level scale selection. 
Despite the significant improvements brought by these methods, the essential problem of insufficient patch receptive fields remains unresolved, leading to suboptimal reconstruction in textureless areas.

\begin{figure*}
\centering
\includegraphics[width=\linewidth]{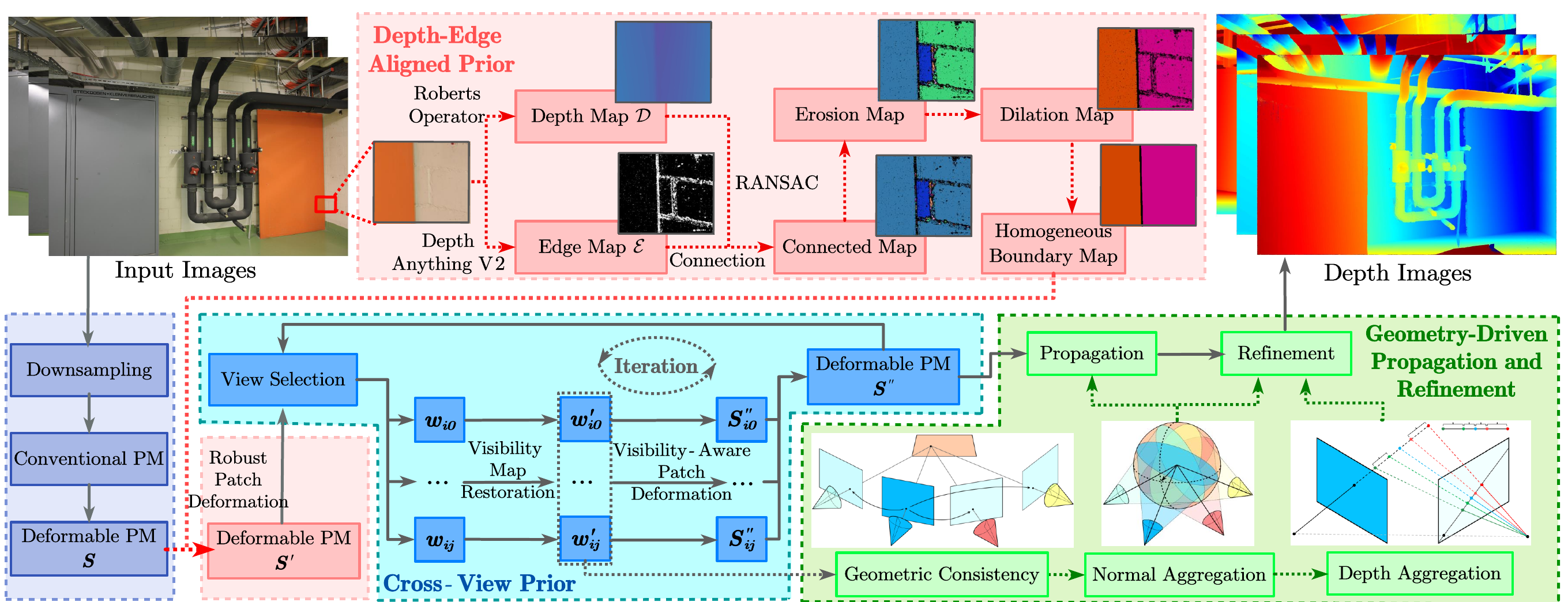}
\caption{Pipeline of DVP-MVS. We first adopt Depth Anything V2 followed by the Roberts operator to initialize corresponding depth and edge maps, respectively. We then employ an erosion-dilation strategy to extract the \textbf{depth-edge aligned prior} for robust patch deformation. Subsequently, we construct visibility maps by reforming view selection and adopting the reprojection-based post-verification for visibility map restoration, which are then treated as the \textbf{cross-view prior} to facilitate visibility-aware patch deformation. Finally, by considering geometric consistency, we respectively improve the propagation and refinement stages by introducing visible normals aggregation and epipolar line projection.
After several iterations we obtain depth images.
}
\label{fig:pipeline}
\end{figure*}

\subsubsection{Patch Deformation. } As a branch of PatchMatch MVS, patch deformation adaptively expands patch receptive fields to help reconstruct textureless areas.
Both patch resizing and patch reshaping belongs to patch deformation. For patch resizing, PHI-MVS \cite{PHI-MVS} and API-MVS \cite{API-MVS} respectively introduce dilated convolution and entropy calculation to modify patch size and sampling interval. However, an overlarge interval may cause sparse patch attention. For patch reshaping, SD-MVS \cite{SD-MVS} employs instance segmentation to constrain patch deformation within instances, while its results heavily depend on actual segmentation quality.
Moreover, APD-MVS \cite{APD-MVS} separates the unreliable pixel's patch into several outward-spreading sub-patches with high reliability. 
However, without depth edge constraint, deformed patches may occur edge-skipping to cover areas with depth-discontinuity.

\subsection{Learning-based MVS Method}
The development of deep learning has led to the advancement of self-supervised learning methods \cite{chen2023self, chen2024learning, DING1, DING2}, parallel computing \cite{Cheng1, Cheng2, Cheng3}, transformer algorithms \cite{SUN1, SUN2}, vision-language model \cite{chen2024bimcv, qianmaskfactory, chen2024tokenunify} and compression strategy \cite{Zhang2, Zhang1, Zhang3}.
In MVS field, MVSNet \cite{MVSNet} pioneers the use of neural networks to construct 3D differentiable cost volumes for reconstruction.
IterMVS-LS\cite{IterMVS} further incorporate the GRU module for regularization to solve the reconstruction of high-resolution scenes.
Cas-MVSNet \cite{Cas-MVSNet} leverage a coarse-to-fine strategy to achieve progressively refined depth estimation.
MVSTER \cite{MVSTER} proposes an epipolar transformer to learn both 2D semantics and 3D spatial associations. 
EPP-MVSNET \cite{EPP-MVSNet} introduces an epipolar-assembling module to package images into limited cost volumes. 
GeoMVSNet \cite{GeoMVSNet} integrates geometric priors from coarse stages through a two-branch fusion network to refine depth estimation.
RA-MVSNet \cite{RA-MVSNet} utilizes a two-branch framework that integrates distance integration within cost volume to reconstruct textureless area.
Despite this, most methods suffer from large training datasets, unaffordable memory usage or limited generalization.

\section{Method}
\subsection{Overview}
Given a series of input images $\mathcal{I} = \{I_i | i = 1, ..., N \}$ and their corresponding camera parameters $\mathcal{P}=\{K_i, R_i, T_i \mid i=1 \cdots N\}$, we sequentially select the reference image $I_{\mathrm{ref}}$ from $\mathcal{I}$ and reconstruct its depth map through pairwise matching with the remaining source images $I_{\mathrm{src}}(\mathcal{I} - I_{\mathrm{ref}})$.
Fig. \ref{fig:pipeline} shows our method's pipeline, and the implementation of each component will be detailed in the following sections. 

\begin{figure*}
\centering
\includegraphics[width=\linewidth]{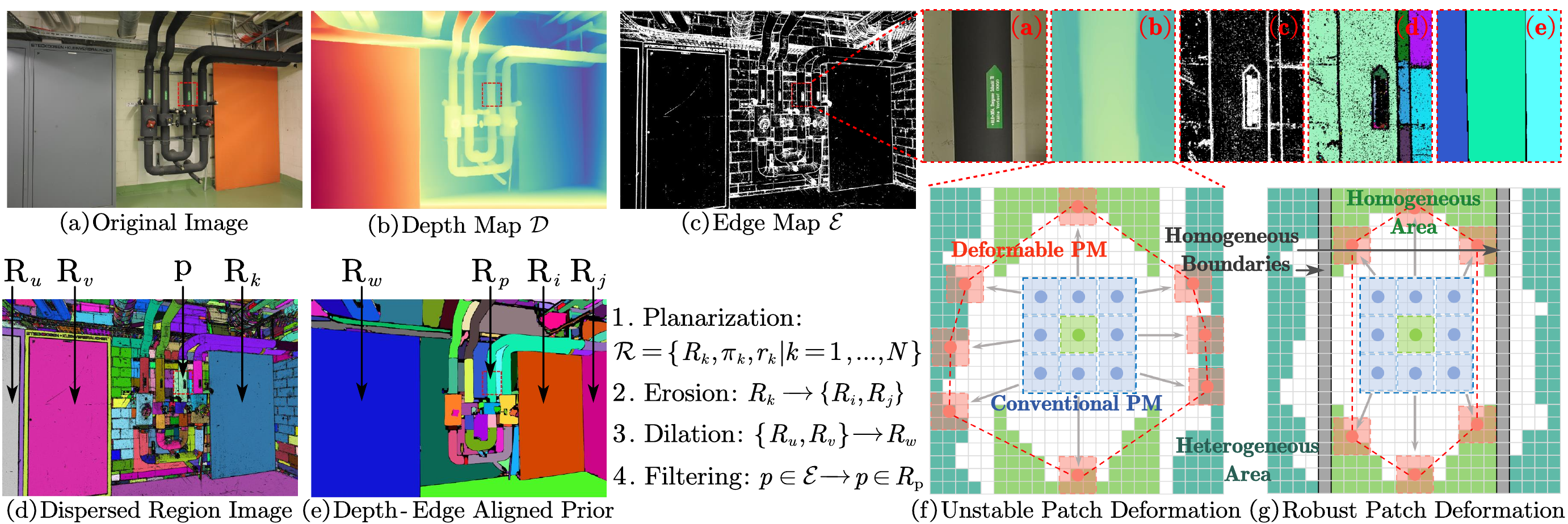}
\caption{
Depth-Edge Aligned Prior. Edges are highlighted in white in (c), with black constituting dispersed regions.
Different colors denote different dispersed regions in (d) and homogeneous areas in (e), with black indicating their boundaries. In (f) and (g), green, blue and red respectively denote the central pixel, neighbors in conventional PM and anchors in deformable PM. 
Cyan, green and gray backgrounds respectively denote heterogeneous areas, homogeneous areas and homogeneous boundaries.
}
\label{fig: DEP}
\end{figure*}




\subsection{Preliminary Area}
The conventional PM method projects a fixed-size patch in the reference image onto a corresponding patch in the source image with a given plane hypothesis.
Subsequently, it calculates the similarity between the two patches using the NCC matrix~\cite{COLMAP} for evaluation.
Specifically, given the reference image $I_i$ and the source image $I_j$, for each pixel $p$ in $I_i$, we first randomly generate a plane hypothesis $(\mathbf{n}^T, d)$, where $\textbf{n}$ and $d$ respectively denote the normal and the depth. Through homography mapping \cite{Accurate}, we can calculate the projection matrix $H_{ij}$ of the plane hypothesis $(\mathbf{n}^T, d)$ for pixel $p$ between $I_i$ and $I_j$. 
We then employ $H_{ij}$ to project the fixed-size patch $B_p$ centered at $p$ in the reference image $I_i$ onto the mapping patch $B_p^j$ in the source image $I_j$. Consequently, the matching cost is calculated as the NCC score between $B_p$ and $B_p^j$, formulated by: 
\begin{equation}
m_{ij}\left(p, B_p\right)=1-\frac{cov\left(B_p, B_p^j\right)}{\sqrt{cov\left(B_p, B_p\right) cov\left(B_p^j, B_p^j\right)}},
\end{equation}
where $cov$ is weighted covariance \cite{COLMAP}. 
Additionally, the multi-view aggregated cost is defined as:
\begin{equation}
m\left(p, B_p\right)=\frac{\sum^{N-1}_{j=1} w_{ij}(p) \cdot m_{ij}\left(p, B_p\right)}{\sum^{N-1}_{j=1} w_{ij}(p)},
\end{equation}
where $w_{ij}(p)$ is gained from the view selection strategy~\cite{ACMMP}. 
Finally, through propagation and refinement, we bring multiple hypotheses to each pixel for cost computation and select the minimal-cost hypothesis as the final result.


In contrast, the deformable PM \cite{APD-MVS} decomposes the patch of the unreliable pixel into several outward-spreading, reliable sub-patches, with the center of each sub-patch regarded as the \textbf{anchor} pixel.
Specifically, for each unreliable pixel $p$, its matching cost of deformable PM is: 
\begin{equation}
m_{ij}(p, S) = \lambda m_{ij}(p, B_p) + (1 - \lambda) \frac{\sum_{s \in S} m_{ij}(s, B_s)}{|S|},
\label{PM}
\end{equation}
where $S$ denotes the collection of all anchors, $B_s$ represents the sub-patch with each anchor pixel $s \in S$ as center.
In practice, the patch sizes of $B_p$ and $B_s$ are both $11 \times 11$, while respective have sampling intervals 5 and 2.
In experiment, $\lambda = 0.25, |S| = 8$.

\subsection{Depth-Edge Aligned Prior}
To reconstruct unreliable pixels in textureless areas, recent methods adopt patch deformation to expand the patch receptive field, while ignoring the critical depth edge constraint on deformation stability. 
As shown in Fig.~\ref{fig: DEP} (f), the absence of depth edge constraint causes the deformed patch to occur edge-skipping and cover heterogeneous areas with depth discontinuity, leading to potential estimation deviation.

To fully exploit depth edges, we introduce the depth-edge aligned prior for robust patch deformation. Given image in Fig.~\ref{fig: DEP} (a), we first utilize Depth Anything V2~\cite{depany2} to initialize \textbf{depth} maps $\mathcal{D}$ in Fig.~\ref{fig: DEP} (b). 
Depth Anything V2 is a robust monocular depth estimation model that can achieve zero-shot depth generalization across diverse scenarios.
Meanwhile, we use Roberts operator on image $I$ to extract coarse edge maps $\mathcal{D}$ and dispersed regions $\mathcal{R}$ by connecting the remaining non-edge areas, represented with different colors in Fig.~\ref{fig: DEP} (d). The whole process can be defined by: $I=\mathcal{D} \cup \mathcal{R}$. 
Moreover, we respectively define the heterogeneous region $H$, the homogeneous region $M$ and the homogeneous boundary $E$ as follows: $H=\{p \in \Omega \mid \nabla d_p > \epsilon \}$, $M=\{p \in \Omega \mid \nabla d_p \leq \epsilon \}$, $E_{i,j}=M_i \cap M_j$, where $\nabla d_p$ denotes depth gradient of pixel $p$, $\epsilon$ is set to 0.005.
In addition, to align depth and edge maps for the generation of fine-grained homogeneous boundaries, a region-level erosion-dilation strategy is proposed to distinguish and aggregate homogeneous areas. 

Specifically, given the dispersed region image shown in Fig.~\ref{fig: DEP} (d), for each it dispersed region $R_k$ whose size exceeds $\eta$, we combine depth and region information from $\left[\mathcal{D},\mathcal{E}\right]$ to perform region-wise planarization with RANSAC, acquiring estimated plane $\pi_k = (n_k, d_k)$.
Additionally, we define the ratio of inlier pixels to total pixels as the inlier ratio $r_k$.
Therefore, we obtain $\mathcal{R} = \{R_k, \pi_k, r_k | k = 1, ..., N \}$.
During the erosion stage, we perform intra-regional erosion to separate regions belonging to heterogeneous areas.
For instance, region $R_k$ is pre-divided into sub-regions $R_i$ and $R_j$ by erosion.
We then reapply planarization on both sub-regions to respectively acquire estimated planes $\pi_i=(n_i, d_i)$, $\pi_j=(n_j, d_j)$ and inlier ratio $r_i$, $r_j$. 
Ultimately, we consider erosion is effective and divide region $R_k$ when:
\begin{equation}
R_k \xrightarrow{}\left\{R_i, R_j\right\} \text {, if } \text{sim}(\pi_i, \pi_j) \leq \sigma \text { s.t. } \frac{r_i+r_j}{2 r_k} \geq \gamma ,
\label{erosion}
\end{equation}
where plane similarity function $\text{sim}(\pi_i, \pi_j)$ is defined by: 
\begin{equation}
\text{sim}(\pi_i, \pi_j) = \mathbf{n}_i \cdot \mathbf{n}_j + \min(1, |d_i - d_j|),
\end{equation}
in Eq. \ref{erosion}, a higher inliers ratio indicates a superior planarization result after erosion, and a lower similarity in estimated planes indicates both sub-regions are located in homogeneous areas. 
On the contrary, for the dilation stage, we conduct inter-regional dilation to merge regions belonging to homogeneous areas. For example, regions $R_u$ and $R_v$ are pre-merged into an aggregated region $R_w$ through dilation. We then distinguish dilation is valid and perform region mergence under the following condition:
\begin{equation}
\{R_u, R_v\} \xrightarrow{\text{}} R_w\text{, if } \text{sim}(\pi_u, \pi_v) \geq \sigma \text{ s.t. } r_u, r_v \geq \kappa,
\label{dilation}
\end{equation}
in Eq. \ref{dilation}, if two estimated planes are both reliable and similar to each other, we consider regions $R_u$ and $R_v$ should be merged together as homogeneous areas. 

Following erosion-dilation strategy, we further incorporate pixel-wise filtering to refine fine-grained homogeneous boundaries. Specifically, when the pixel $p=(x, y, d_p)$ with depth $d_p$ is adjacent to region $R_p$ with estimated plane $\pi_k = (\mathbf{n}_k, d_k)$, we regard $p$ belonging to $R_p$ provided that:
\begin{equation}
p \in \mathcal{E} \xrightarrow{\text{}} p \in R_p\text{, if } \frac{| \mathbf{n}_k \cdot p + d_k |}{|\mathbf{n}_k|} \leq \delta \text{ s.t. } r_k \geq \kappa,
\label{filtering}
\end{equation}
in Eq. \ref{filtering}, if the 3D position of $p$ is close to the estimated plane, then $p$ should be considered as a part of region $R_p$.
After obtaining the depth-edge aligned prior, we adopt it to guide patch deformation within homogeneous areas. Specifically, as shown in Fig.~\ref{fig: DEP} (g), for unreliable pixel $p$, we first perform patch deformation to obtain the anchor collection $S$. Then for each anchor $s_i \in S$, we retain it if it lies in the same region as $p$; otherwise we discard it, formulated by:
\begin{equation}
S^{\prime} = \{ s_i \in S \mid s_i \in R_p \},
\label{depth-edge}
\end{equation}
where $S^{\prime}$ indicates the new anchor collection and $R_p$ signifies the homogeneous areas that $p$ belongs to.
Finally, the new collection $S^{\prime}$ will substitute the old one $S$ for patch deformation, thereby avoiding unexpected edge-skipping and guaranteeing deformed patches within homogeneous areas.
 





\subsection{Cross-View Prior}

In addition to the aforementioned edge-skipping, the visibility discrepancy and occlusion caused by viewpoint changes is another crucial issue during patch deformation. 
As shown in Fig.~\ref{fig: view}, partial areas of the reference image in (a) may correspond to invisible areas in the source image in (b). However, the deformed patch in (e) frequently includes these invisible areas for matching cost, leading to potential distortions. 
Therefore, we attempt to progressively integrate cross-view prior to facilitate visibility-aware patch deformation.

\subsubsection{Visibility Map Restoration. }
\begin{figure}
\centering
\includegraphics[width=\linewidth]{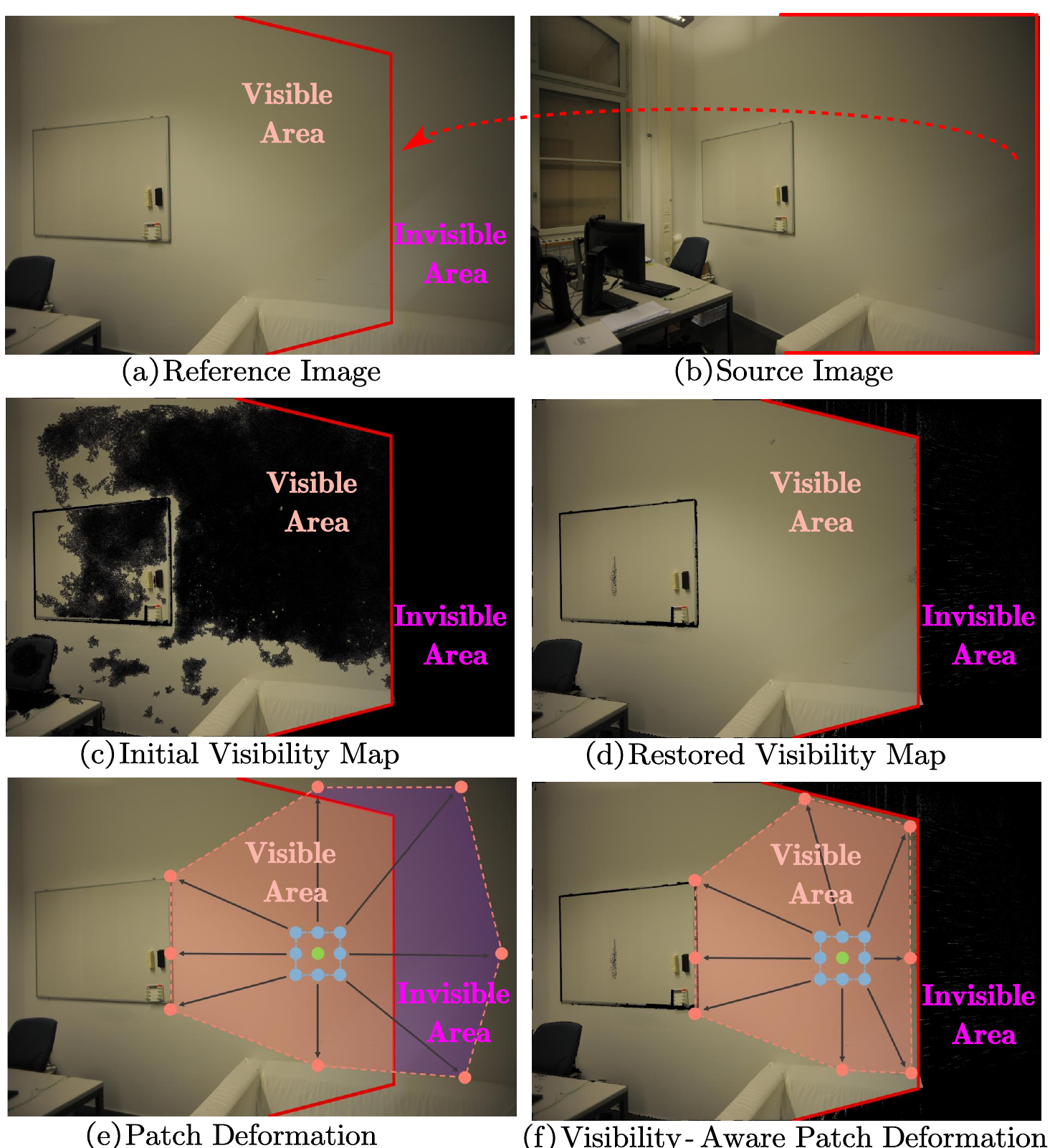}
\caption{Cross-View Prior. 
The red line in (a) separates visible and invisible areas of (a) within (b). In (c) and (d), black indicates pixels judged invisible by view selection strategy. 
In (e) and (f), green, blue and red indicate the central pixel, conventional PM neighbors and deformable PM anchors.
}
\label{fig: view}
\end{figure}

For each pixel $p$ in the reference image $I_i$, we first initialize its visibility weights $w_{ij}(p)$ corresponding to source images $I_j$ through view selection strategy~\cite{ACMMP}.
Here $w_{ij}(p)$ is not simply used for cost aggregation but also reformulated as visibility maps to provide patch deformation with cross-view visibility perception.
According to \cite{ACMMP}, $w_{ij}(p)$ is determined by evaluating costs against fixed thresholds. Therefore, low costs are more likely to be visible, while high costs are often deemed invisible.
Thus, in textureless areas characterized by high costs, original-visible areas are often mistakenly judged as invisible, as shown in Fig.\ref{fig: view} (c). Therefore, visibility determined merely by cost is unreliable, as it struggles to strike the balance between well-textured and textureless areas. 

Essentially, visibility signifies the presence of corresponding pixel pairs between images, meaning each pixel can be mapped to the other via depth projection. Thus, we subtly adopt the cross-view depth reprojection $e(p)$~\cite{ACMMP} as post-verification for visibility map restoration. Through back-and-forth projection, $e(p)$ can effectively validate the visibility between pixel pairs and the robustness of depths.

Specifically, for each pixel $p$ with depth $d$ in invisible areas of the reference image $I_i$, we first project it into source image $I_j$ to obtain pixel $p_j$. We then reproject $p_j$ with its corresponding depth $d_j$ in $I_j$ back into $I_i$ to derive reprojection pixel $p^{\prime}$. Ultimately, we define $e(p)=\left\|p^{\prime}-p\right\|$ and regard $p$ as visible with respect to $I_j$ when $e(p) \leq \varepsilon$, thereby generating restored weight $w_{ij}^{\prime}(p)$ and visibility map in Fig.~\ref{fig: view} (d), which effectively restores the visibility in textureless areas.

\begin{figure*}
\centering
\includegraphics[width=\linewidth]{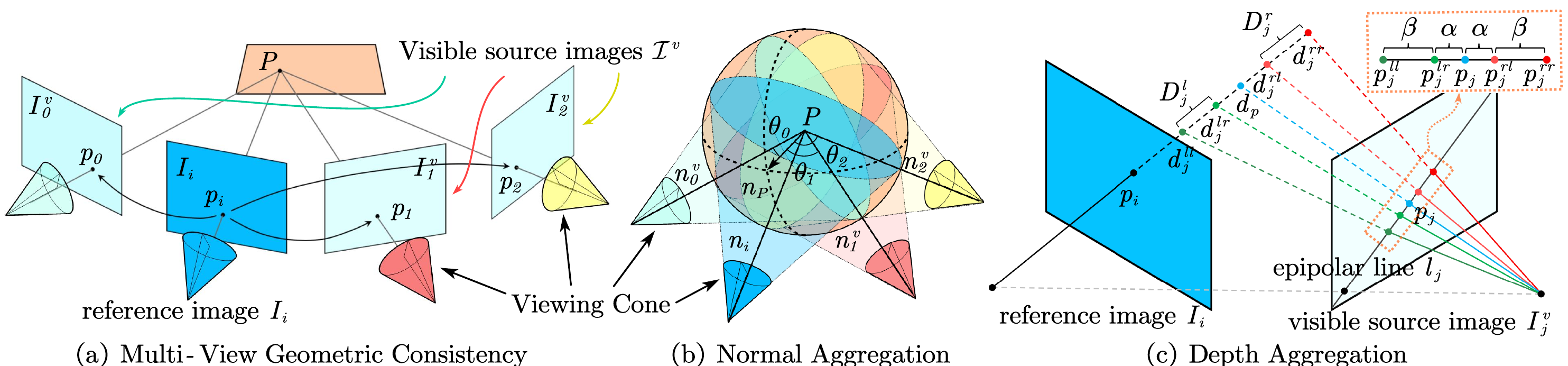}
\caption{Geometry-Driven Propagation and Refinement. In (a) and (b), the blue view cone corresponds to the reference image $I_i$, while green, red, and yellow view cones correspond to visible source images $I^v_0$, $I^v_1$ and $I^v_2$. 
}
\label{fig: epip}
\end{figure*}

\subsubsection{Visibility-Aware Patch Deformation. }
After acquiring the restored visibility map, we further regard it as cross-view prior to iteratively update patch deformation. 
Specifically, for each unreliable pixel $p$ in the reference image $I_i$, we first perform patch deformation to obtain anchor collection $S$, which is then guided by depth-edge prior to obtain $S^{\prime}$ via Eq. \ref{depth-edge}. Then for each anchor $s_i \in S^{\prime}$, we retain it if it is visible in source image $I_j$; otherwise, we discard it. Combined with Eq. \ref{depth-edge}, the view-level anchor collection $S_{ij}^{\prime}$ is defined by: 
\begin{equation}
S_{ij}^{\prime\prime} = \{ s_i \in S^{\prime} \mid s_i \in R_p, w_{ij}(p) > 0 \}.
\end{equation}
Subsequently, $S_{ij}^{\prime}$ will substitute $S$ in Eq. \ref{PM} to calculate the matching cost $m_{ij}(p, S_{ij}^{\prime})$, which in turn further optimizes $w_{ij}^{\prime}(p)$ during view selection, thereby iteratively facilitating visibility-aware patch deformation, as shown in Fig.~\ref{fig: view} (f). Finally, the final multi-view aggregated cost $m(p,S^{\prime\prime})$ is:
\begin{equation}
m\left(p, S^{\prime\prime}\right)=\frac{\sum^{N-1}_{j=1} w_{ij}^{\prime}(p) \cdot m_{ij}\left(p, S_{ij}^{\prime\prime}\right)}{\sum^{N-1}_{j=1} w_{ij}^{\prime}(p)}.
\end{equation}




%

\subsection{Geometry-Driven Propagation and Refinement}
Previous methods \cite{ACMMP} incorporate neighborhood optimal hypotheses (i.e., normals and depths) for propagation and generate random hypotheses for refinement. 
However, the potential geometric constraint may render partial hypotheses invalid.
Addressing this, we leverage the multi-view geometric consistency to improve propagation and refinement stages, including \textbf{normal} and \textbf{depth} aggregations.

\subsubsection{Normal Aggregation.}
To ensure geometric consistency, we first adopt the restored visibility maps to aggregate multi-view normals, thus restraining the normals in visible ranges.
As shown in Fig. \ref{fig: epip} (a), for each pixel $p_i$ in reference image $I_i$, we first leverage its restored visibility maps to identify all visible source images $\mathcal{I}^v = \{I^v_j | j = 1, ..., N - 1, w_{ij}^{\prime}(p) > 0\}$. We then utilize depth projection to obtain the 3D point $P$ and mapping pixels $p_j$ for each visible source image $I^v_j$. 
Since each $I^v_j$ in $\mathcal{I}^v$ has a visible hemisphere within its viewing cone and $P$ is visible to $I^v_j$, we can aggregate all visible hemispheres to constrain the normal range of 3D point $P$.

Specifically, we acquire the viewing cone direction $n^v_j$ of each visible source image $I^v_j$ by connecting its camera center with the mapping pixels $p_j$.
Then, for each $I^v_j$, we consider the angle $\theta_j$ between its viewing cone direction $n^v_j$ and the normal $n_P$ of $P$ should exceed $90^{\circ}$ (i.e., $n_P \cdot n^v_j \leq 0$), as shown in Fig. \ref{fig: epip} (b).
Finally, by reprojecting the normal range of $P$ back to pixel $p_i$ in $I_i$, we effectively constrain the normal range of $n_i$. 
The aggregated normal range then serves as the visibility constraint during propagation and refinement, thereby enhancing the reliability of normal hypotheses.

\subsubsection{Depth Aggregation. }
For depth hypotheses, there exists two primary problems for conventional fixed depth intervals during the refinement process: $1)$ when depth interval is uniformly distributed in $I_{\mathrm{ref}}$ and projected onto $I_{\mathrm{src}}$, the spacing of the corresponding points along the epipolar line becomes uneven and progressively narrower as depth increases; $2)$ for each depth interval in $I_{\mathrm{ref}}$, the projected intervals along the epipolar lines can be variable in different source images.

Therefore, we perform inverse projection and aggregation of fixed-length sampling along multi-view epipolar lines to finetune the depth interval. 
Specifically, as illustrated in Fig. \ref{fig: epip} (c), for each pixel $p_i$ in reference image $I_i$, we first identify its mapping pixel $p_j$ along epipolar line $l_j$ in visible source image $I^v_j$ as previously mentioned. 
We then select four pixels, located at distances $\alpha$ and $\alpha + \beta$ from $p$ on either side along epipolar line $l_j$, as the minimum and maximum disturbance extremes of mapping pixels. 
These four pixels are further inversely projected into the reference image $I_i$ to obtain their corresponding depths, thus forming two intervals for refinement:
$D^l_{j} = (d^{ll}_{j}, d^{lr}_{j})$ and $D^r_{j} = (d^{rl}_{j}, d^{rr}_{j})$. 
Such adaptive intervals ensure that the mapping pixel of refined hypothesis can achieve appropriate displacements along the epipolar line without being constrained by the current depth. 

\begin{figure*}
\centering
\includegraphics[width=\linewidth]{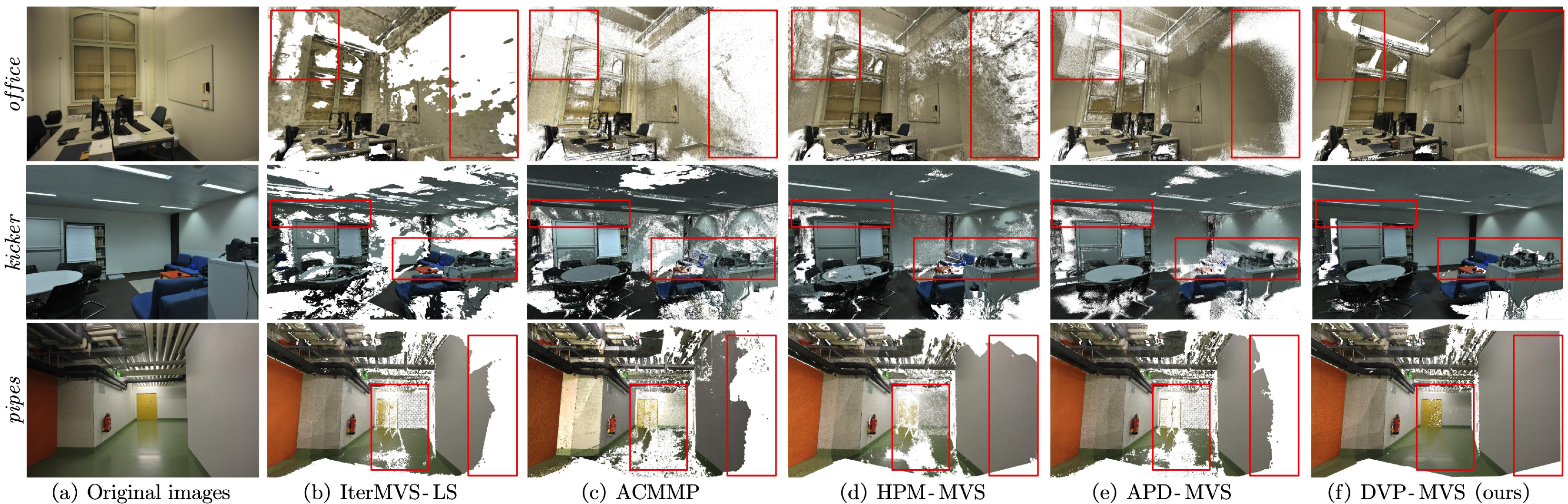}
\caption{Qualitative results on partial scenes of ETH3D datasets (\emph{office}, \emph{kicker} and \emph{pipes}). Our method can achieve the highest completeness while containing fewer outliers, especially in large textureless areas like walls and floors, as shown in red boxes.
}
\label{fig: eth3d results}
\end{figure*}

Moreover, to ensure mapping pixels experiences notably displacements in as many source images as possible, we aggregate depth intervals across all visible source images $\mathcal{I}^v$. Specifically, we identify the extremes of aggregated intervals as the $\mu^{th}$ smallest value among all maximum disturbance extremes (i.e., $d^{ll}_{j}$ and $d^{rr}_{j}$) and the $\mu^{th}$ largest value among all minimum disturbance extremes (i.e., $d^{lr}_{j}$ and $d^{rl}_{j}$):
\begin{equation}
\left(\min_{1\leq j\leq \mu} d^{ll}_{j}, \max_{1\leq j\leq \mu} d^{lr}_{j}\right), 
\left(\max_{1\leq j\leq \mu} d^{rl}_{j}, \min _{1\leq j\leq \mu} d^{rr}_{j}\right).
\end{equation}
A detailed explanation of this equation is shown in supplementary material.
Finally, the aggregated depth interval will replace the fixed depth interval during local perturbations of refinement to enhance the reliability of depth hypotheses.

\section{Experiment}
\subsection{Datasets, Metrics and Implementation Details}
We evaluate our work on both ETH3D \cite{ETH3D} and Tanks \& Temples (TNT) \cite{TNT} datasets and upload our results to their websites for reference. 
We compare our work against state-of-the-art learning-based methods like PatchMatchNet, IterMVS, MVSTER, AA-RMVSNet, EPP-MVSNet, EPNet and state-of-the-art traditional MVS methods like TAPA-MVS, PCF-MVS, ACMM, ACMP, ACMMP, SD-MVS, APD-MVS and HPM-MVS.

Concerning metrics, we adopt the percentage of accuracy and completeness for evaluation, with the F$_1$ score adopted as their harmonic mean to measure the overall quality.

We take APD-MVS \cite{APD-MVS} as our baseline. Concerning parameter setting, $\{\eta, \sigma, \gamma, \kappa, \delta, \varepsilon, \alpha, \beta, \mu\} = \{3 \times 10^{2}, 0.5, 1.2, 0.7, 0.8, 2, 1, 4, 3\}$. 

Our method is implemented on a machine with an Intel(R) Xeon(R) Silver 4210 CPU and eight NVIDIA GeForce RTX 3090 GPUs. 
We take APD-MVS \cite{APD-MVS} as our baseline. Experiments are performed on original images in both ETH3D and TNT datasets. In cost calculation, we adopt the matching strategy of every other row and column. 




\begin{table}
  \centering
  \renewcommand{\arraystretch}{1.05} 
    \resizebox{\linewidth}{!}{
        \begin{tabular}{c|ccc|ccc}
        \hline
        \multirow{2}{*}{Method} & \multicolumn{3}{c|}{Train} & \multicolumn{3}{c}{Test} \\
        \cline{2-7} & F$_1$ & Comp. & Acc. & F$_1$ & Comp. & Acc. \\   
        \hline 
        \multirow{1}{*}{PatchMatchNet} & 64.21 & 65.43 & 64.81 & 73.12 & 77.46 & 69.71 \\  
        \multirow{1}{*}{IterMVS-LS} & 71.69 & 66.08 & 79.79 & 80.06 & 76.49 & 84.73 \\  
        \multirow{1}{*}{MVSTER} & 72.06 & 76.92 & 68.08 & 79.01 & 82.47 & 77.09 \\  
        \multirow{1}{*}{EPP-MVSNet} & 74.00 & 67.58 & 82.76 & 83.40 & 81.79 & 85.47 \\  
        \multirow{1}{*}{EPNet} & 79.08 & 79.28 & 79.36 & 83.72 & 87.84 & 80.37 \\  
        \hline 
        \multirow{1}{*}{TAPA-MVS} & 77.69 & 71.45 & 85.88 & 79.15 & 74.94 & 85.71 \\  
        \multirow{1}{*}{PCF-MVS} & 79.42 & 75.73 & 84.11 & 80.38 & 79.29 & 82.15 \\  
        \multirow{1}{*}{ACMM} & 78.86 & 70.42 & \textbf{90.67} & 80.78 & 74.34 & 90.65 \\  
        \multirow{1}{*}{ACMMP} & 83.42 & 77.61 & \underline{90.63} & 85.89 & 81.49 & \textbf{91.91} \\  
        \multirow{1}{*}{SD-MVS} & 86.94 & 84.52 & 89.63 & 88.06 & 87.49 & 88.96 \\ 
        \multirow{1}{*}{HPM-MVS++} & \underline{87.09} & \underline{85.64} & 88.74 & \underline{89.02} & \underline{89.37} & 88.93 \\  
        \multirow{1}{*}{APD-MVS (base)} & 86.84 & 84.83 & 89.14 & 87.44 & 85.93 & 89.54 \\  
        \hline 
        \multirow{1}{*}{DVP-MVS (ours)} & \textbf{88.67} & \textbf{88.07} & 89.40 & \textbf{89.60} & \textbf{89.41} & \underline{91.32} \\  
        \hline
        \end{tabular}%
    }
  \caption{Quantitative results on ETH3D dataset at threshold $2cm$. Our method achieves the highest F$_1$ and completeness.}
  \label{table: eth3d results}%
\end{table}%

\subsection{Benchmark Performance}
Regarding ETH3D dataset, Fig. \ref{fig: eth3d results} shows the qualitative comparison. Our method can reconstruct textureless areas without detail distortion, as shown in red boxes.
Tab. \ref{table: eth3d results} exhibits the quantitative analysis, where the first group is learning-based methods and the second is traditional methods. Meanwhile, the best and the second-best results are respectively marked in bold and underlined. Our method achieves the highest F$_1$ scores and completeness on both training and testing datasets, validating its excellent effectiveness.

Regarding TNT dataset, we test our method on it \textbf{without fine-tuning} to prove generalization. Tab. \ref{table: TNT results} exhibits the quantitative analysis. Our method achieves the highest completeness on both advanced and intermediate datasets and the highest F$_1$ scores on intermediate dataset, demonstrating its astonishing robustness. 
More qualitative results on both ETH3D and TNT datasets along with a detailed memory and runtime comparison are shown in supplementary materials.

\begin{table}
    \centering
    \renewcommand{\arraystretch}{1.05} 
    \resizebox{\linewidth}{!}{
        \begin{tabular}{c|ccc|ccc}
        \hline
        \multirow{2}{*}{Method} & \multicolumn{3}{c|}{Intermediate} & \multicolumn{3}{c}{Advanced} \\
        \cline{2-7} & F$_1$ & Rec. & Pre. & F$_1$ & Rec. & Pre. \\   
        \hline 
        \multirow{1}{*}{PatchMatchNet} & 53.15 & 69.37 & 43.64 & 32.31 & 41.66 & 27.27 \\  
        \multirow{1}{*}{AA-RMVSNet} & 61.51 & 75.69 & 52.68 & 33.53 & 33.01 & 37.46 \\  
        \multirow{1}{*}{IterMVS-LS} & 56.94 & 74.69 & 47.53 & 34.17 & 44.19 & 28.70 \\  
        \multirow{1}{*}{MVSTER} & 60.92 & \underline{77.50} & 50.17 & 37.53 & 45.90 & 33.23 \\  
        \multirow{1}{*}{EPP-MVSNet} & 61.68 & 75.58 & 53.09 & 35.72 & 34.63 & \underline{40.09} \\
        \multirow{1}{*}{EPNet} & \underline{63.68} & 72.57 & \underline{57.01} & \textbf{40.52} & \underline{50.54} & 34.26 \\ 
        \hline 
        \multirow{1}{*}{ACMM} & 57.27 & 70.85 & 49.19 & 34.02 & 34.90 & 35.63 \\  
        \multirow{1}{*}{ACMP} & 58.41 & 73.85 & 49.06 & 37.44 & 42.48 & 34.57 \\  
        \multirow{1}{*}{ACMMP} & 59.38 & 68.50 & 53.28 & 37.84 & 44.64 & 33.79 \\  
        \multirow{1}{*}{SD-MVS} & 63.31 & 76.63 & 53.78 & 40.18 & 47.37 & 35.53 \\  
        \multirow{1}{*}{HPM-MVS++} & 61.59 & 73.79 & 54.01 & 39.65 & 41.09 & \textbf{40.79} \\  
        \multirow{1}{*}{APD-MVS (base)} & 63.64 & 75.06 & \underline{55.58} & 39.91 & 49.41 & 33.77 \\  
        \hline 
        \multirow{1}{*}{DVP-MVS (ours)} & \textbf{64.76} & \textbf{78.69} & 55.04 & \underline{40.23} & \textbf{54.21} & 32.16 \\  
        \hline
        \end{tabular}%
    }
    \caption{Quantitative results on TNT dataset at given threshold. Our method achieves the highest F$_1$ and completeness.}
    \label{table: TNT results}%
\end{table}%

\subsection{Memory \& Runtime Comparison}
\begin{figure}
    \centering
    \includegraphics[width=\linewidth]{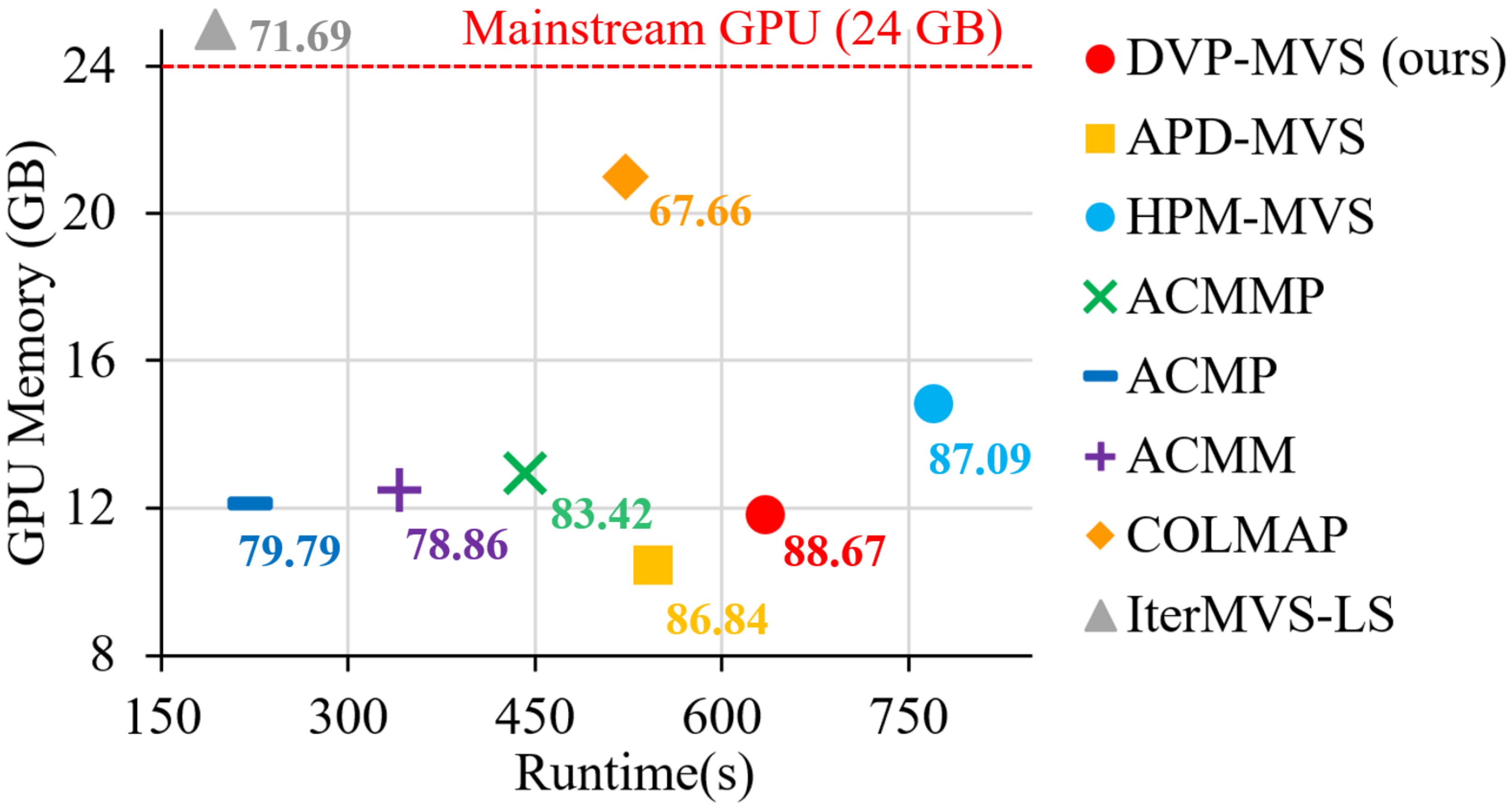}
    \caption{GPU memory usage (GB) and runtime (second) between different methods on ETH3D training datasets. 
    }
    \label{fig:mem and time}
\end{figure}

To validate the efficiency of our proposed method, we perform a comparative analysis of GPU memory usage and runtime of different methods on the ETH3D training datasets, as depicted in Fig. \ref{fig:mem and time}. All experiments were conducted on original-resolution images, with the number of images standardized to 10 for consistency in the runtime evaluation. Furthermore, to ensure fairness, all methods were evaluated on the same hardware, whose configuration has been detailed in the previous section. The F$_1$ scores of each method at threshold $2cm$ on the ETH3D training dataset are indicated next to the corresponding pattern for comparsion.

Regarding learning-based MVS methods, IterMVS-LS achieves the shortest runtime but exhibits excessive memory consumption, exceeding the limitations of mainstream GPUs. Other learning-based MVS methods also encounter similar memory-related limitations, thus restricting their applicability for large-scale scenario reconstructions.

Concerning traditional MVS methods, although ACMM, ACMP, and ACMMP achieve relatively shorter runtimes with comparable memory usage, our method delivers significantly better results. Compared to baseline APD-MVS, our method consumes similar runtime and memory usage while achieving superior performance.  Additionally, our method comprehensively outclasses HPM-MVS in terms of runtime, memory and reconstruction quality. In summary, our method can deliver SOTA performance with acceptable runtime and memory usage, demonstrating its considerable practicality.

\begin{table}
    \centering
    \renewcommand{\arraystretch}{1.05} 
    \resizebox{\linewidth}{!}{
        \begin{tabular}{c|ccc|ccc}
        \hline
        \multirow{2}{*}{Method} & \multicolumn{3}{c|}{$2cm$} & \multicolumn{3}{c}{$10cm$} \\
        \cline{2-7} & F$_1$ & Comp. & Acc. & F$_1$ & Comp. & Acc. \\   
        \hline  
        \multirow{1}{*}{w/o. Agn.} & 86.53 & 84.43 & 88.85 & 96.46 & 95.10 & 97.89 \\ 
        \multirow{1}{*}{w/o. Ero.} & 88.02 & 86.95 & 89.23 & 97.52 & 96.89 & 98.15 \\ 
        \multirow{1}{*}{w/o. Dil.} & 87.56 & 86.19 & 89.03 & 97.14 & 96.28 & 98.01 \\ 
        \multirow{1}{*}{w/o. Fil.} & 87.75 & 86.52 & 89.12 & 97.28 & 96.51 & 98.07 \\ 
        \hline
        \multirow{1}{*}{w/o. Cro.} & 86.76 & 84.92 & 88.79 & 96.61 & 95.43 & 97.85 \\ 
        \multirow{1}{*}{w/o. Res.} & 87.84 & 86.67 & 89.15 & 97.35 & 96.64 & 98.10 \\ 
        \multirow{1}{*}{w/o. Vis.} & 87.39 & 86.09 & 88.86 & 97.05 & 96.27 & 97.87 \\ 
        \hline
        \multirow{1}{*}{w/o. Geo.} & 86.92 & 85.39 & 88.63 & 96.73 & 95.76 & 97.74 \\ 
        \multirow{1}{*}{w/o. Pro.} & 88.13 & 87.25 & 89.16 & 97.54 & 96.99 & 98.08 \\ 
        \multirow{1}{*}{w/o. Ref.} & 87.69 & 86.58 & 88.96 & 97.25 & 96.55 & 97.96 \\ 
        \multirow{1}{*}{w/o. Dep.} & 87.62 & 86.36 & 89.04 & 97.21 & 96.42 & 98.01 \\ 
        \hline
        \multirow{1}{*}{DVP-MVS} & \textbf{88.67} & \textbf{88.07} & \textbf{89.40} & \textbf{97.93} & \textbf{97.61} & \textbf{98.27} \\  
        \hline
        \end{tabular}%
    }
    \caption{Quantitative results of the ablation studies on ETH3D benchmark to validate each proposed component.}
    \label{table: ablation study}%
\end{table}%

\subsection{Ablation Study}
Tab. \ref{table: ablation study} exhibits detailed ablation studies on ETH3D dataset. 

\subsubsection{Depth-Edge Aligned Prior} 
We separately exclude the whole depth-edge aligned prior (w/o. Agn.), region erosion (w/o. Ero.), region dilation (w/o. Dil.), and pixel-wise filtering (w/o. Fil.). 
Obviously, w/o. Agn. produces the worst F$_1$ score, highlighting the significance of edge constraint for patch deformation. 
The F$_1$ score of w/o. Ero. is slightly better than w/o. Dil., meaning region dilation exhibits greater impacts than region erosion. 
Moreover, both w/o. Dil. and w/o. Fil. delivere similar F$_1$ score, yet fell short when compared with DVP-MVS, indicating that region erosion and pixel-wise filtering are equally crucial.

\subsubsection{Cross-View Prior} 
We individually remove the entire cross-view prior (w/o. Cro.), visibility map restoration (w/o. Res.), and visibility-aware patch deformation (w/o. Vis.). 
In comparison, w/o. Cro. yields the lowest F$_1$ score, validating the importance of cross-view prior for visibility-aware patch deformation. Moreover, w/o. Res. attains a higher F$_1$ score than w/o. Vis., emphasizing visibility-aware patch deformation is more crucial than visibility map restoration.

\subsubsection{Geometry-Driven Propagation and Refinement} 
We respectively remove the whole geometry-driven module (w/o. Geo.), normal constraint for propagation (w/o. Pro.) and refinement (w/o. Ref.), and depth constrains for refinement (w/o. Dep.). 
Evidently, without geometric constraint, w/o. Geo. obtains the worst F$_1$ score. 
The F$_1$ score of w/o. Pro. surpasses w/o. Ref., suggesting that the normal constraint is more beneficial to refinement than propagation. 
Moreover, the similar F$_1$ score of w/o. Ref. and w/o. Dep. indicates that normal and depth constraints are equivalently contributive.

\section{Conclusion}
In this paper, we propose DVP-MVS to synergize depth-edge aligned and cross-view prior for robust and visibility-aware patch deformation. 
For robust patch deformation, depth and edge information are aligned with an erosion-dilation strategy to guarantee deformed patches within homogeneous areas. 
For visibility-aware patch deformation, we reform and restore visibility maps and regard them as cross-view prior to bring deformed patches visibility perception. 
Additionally, we incorporate multi-view geometric consistency to improve propagation and refinement with reliable hypotheses.
Experimental results on ETH3D and TNT datasets indicate our method's state-of-the-art performance.

\section{Acknowledgements}
This work was supported in part by the Strategic Priority Research Program of the Chinese Academy of Sciences under Grant No. XDA0450203, in part by the National Natural Science Foundation of China under Grant 62172392, in part by the Innovation Program of Chinese Academy of Agricultural Sciences (Grant No. CAAS-CSSAE-202401 and CAAS-ASTIP-2024-AII), and in part by the Beijing Smart Agriculture Innovation Consortium Project (Grant No. BAIC10-2024).

\newpage

\section{Supplementary Material}

\section{More Implementation Details}

Our method is implemented on a machine with an Intel(R) Xeon(R) Silver 4210 CPU and eight NVIDIA GeForce RTX 3090 GPUs. 
We take APD-MVS \cite{APD-MVS} as our baseline. Experiments are performed on original images in both ETH3D and TNT datasets. In cost calculation, we adopt the matching strategy of every other row and column. 

\section{More Results on ETH3D and TNT datasets}
Fig. \ref{fig:eth3d_supp} shows comparative qualitative results of different approaches on partial scenes from the ETH3D dataset. Additionally, Fig. \ref{fig:tnt_intermediate} and Fig. \ref{fig:tnt_advanced} further provide the evaluations within the intermediate and advanced sets of the Tanks and Temples datasets, respectively. 
The results clearly demonstrate that our approach can achieve superior reconstruction quality among other methods, especially when handling large textureless areas in red boxes, without introducing any conspicuous detail distortion, validating its significant performance and remarkable generalization ability.

\begin{figure*}
\centering
\includegraphics[width=0.8\linewidth]{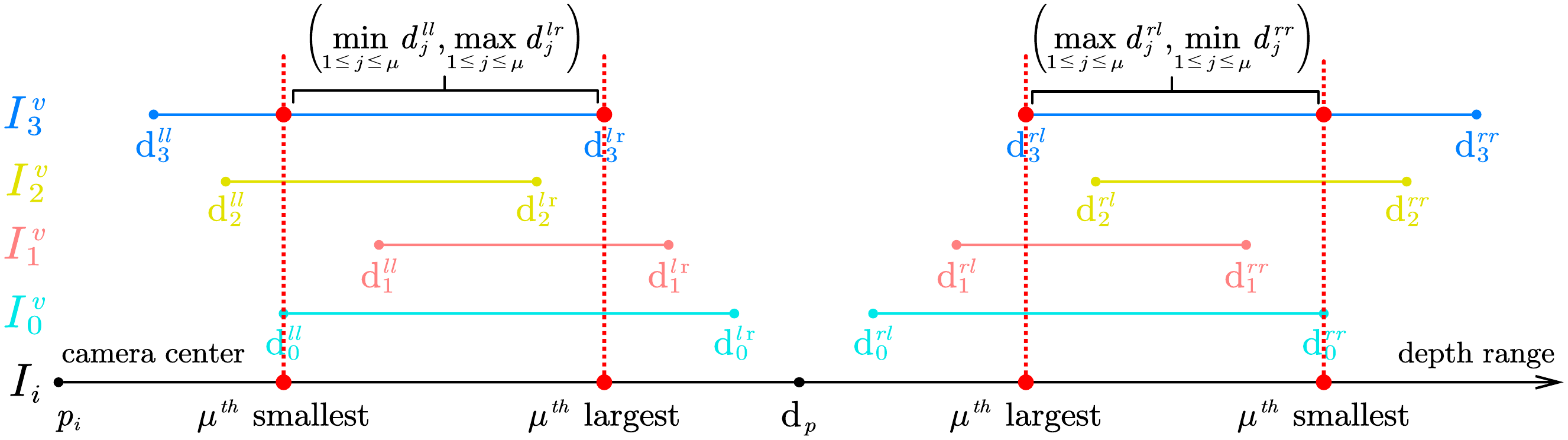}
\caption{Depth interval aggregation. The black line denotes depth range of pixel $p_i$ in reference image $I_i$. Cyan, red, yellow, and blue lines respectively denotes adaptive depth intervals generated from different visible source images $I^v_0$, $I^v_1$, $I^v_2$, and $I^v_3$.
}
\vspace{-0.1in}
\label{fig: depth}
\end{figure*}

\section{Analyses and Discussions}
\textbf{Q1: Why not directly employ the depth discontinuities in the initial depth map obtained from Depth Anything V2 as guidance for patch deformation?}
\par
\noindent\textbf{A1:} 
The core of our method focuses on generating homogeneous boundaries to constrain patch deformation. However, relying on depth discontinuities from the initial depth map produced by Depth Anything V2 can be hard to yield precise homogeneous boundaries as intended. This is because Depth Anything V2 primarily estimates coarse depth information, thereby lacking the fine-grained alignment for accurate depth edge extraction.

Furthermore, when constructing scenes with significant depth variation, applying fixed thresholds for edge extraction is inadequate. In distant areas, depth variations between pixels can be considerable, while in closer areas, these variations are minimal. Therefore, fixed thresholds inevitably fail to accurately delineate edges under such varying conditions, thus resulting in potential inaccuracy.

Therefore, combining the Roberts operator for edge alignment with RANSAC-based planarization to exploit depth information is the optimal strategy for extracting fine-grained homogeneous boundaries for robust patch deformation. 

\par
\noindent\textbf{Q2: Why adopt the smallest value among all maximum disturbance extremes and the largest value among all minimum disturbance extremes for depth aggregation?}
\par
\noindent\textbf{A2:} 
As illustrated in Fig. 1, we depict the adaptive depth intervals of pixel $p$ in the reference image $I_i$ generated from different visible source images $I^v_j$. 
As illustrated in the main text, the adaptive depth intervals  $(d^{ll}_{j}, d^{lr}_{j})$ and $(d^{rl}_{j}, d^{rr}_{j})$ effectively ensures the mapping pixel experiences notably displacements in visible source images $I^v_j$. Additionally, we consider the aggregated intervals should meet such condition in as many source images as possible.

Therefore, we select the $\mu^{th}$ smallest value among all maximum disturbance extremes (i.e., $d^{ll}_j$ and $d^{rr}_j$) and the $\mu^{th}$ largest value among all minimum disturbance extremes (i.e., $d^{lr}_j$ and $d^{rl}_j$) to ensure that the displacement of mapping pixels does not exceed $\alpha + \beta$ and exceeds $\alpha$ on the epipolar line of most visible source images, respectively.



\par
\noindent\textbf{Q3: Why does the original view selection strategy in ACMMP \cite{ACMMP} struggle to balance between well-textured and textureless areas?}
\par
\noindent\textbf{A3:} 
The view selection strategy in ACMMP \cite{ACMMP} determines visibility by assigning matching costs to the current pixel and its neighbors, then comparing these cost values against a fixed threshold. However, due to matching ambiguity, costs in textureless areas is typically unreliable in comparison with those in well-textured areas. 
Consequently, while a fixed threshold can effectively assess the visibility of well-textured areas, it may mistakenly classify originally-visible textureless areas as non-visible, leading in potential inaccuracies.
Thus, the original view selection struggles to balance between well-textured and textureless areas.

\par
\noindent\textbf{Q4: How exactly are the normal and depth constraints applied in the propagation and refinement process?}
\par
\noindent\textbf{A4:} 
During the propagation stage, we select hypothesis with the lowest cost from neighboring pixels in eight directions, ensuring that the selected hypothesis also satisfies the normal constraints. This ensures that the propagated hypotheses are consistent with the expected geometric constraints.

During the refinement stage, we constrain both normal and depth ranges when performing local perturbation or randomization. These constraints ensure that the refined hypotheses remain geometrically consistent, leading to more accurate and reliable depth estimation.

\begin{figure*}
    \centering
    \includegraphics[width=\linewidth]{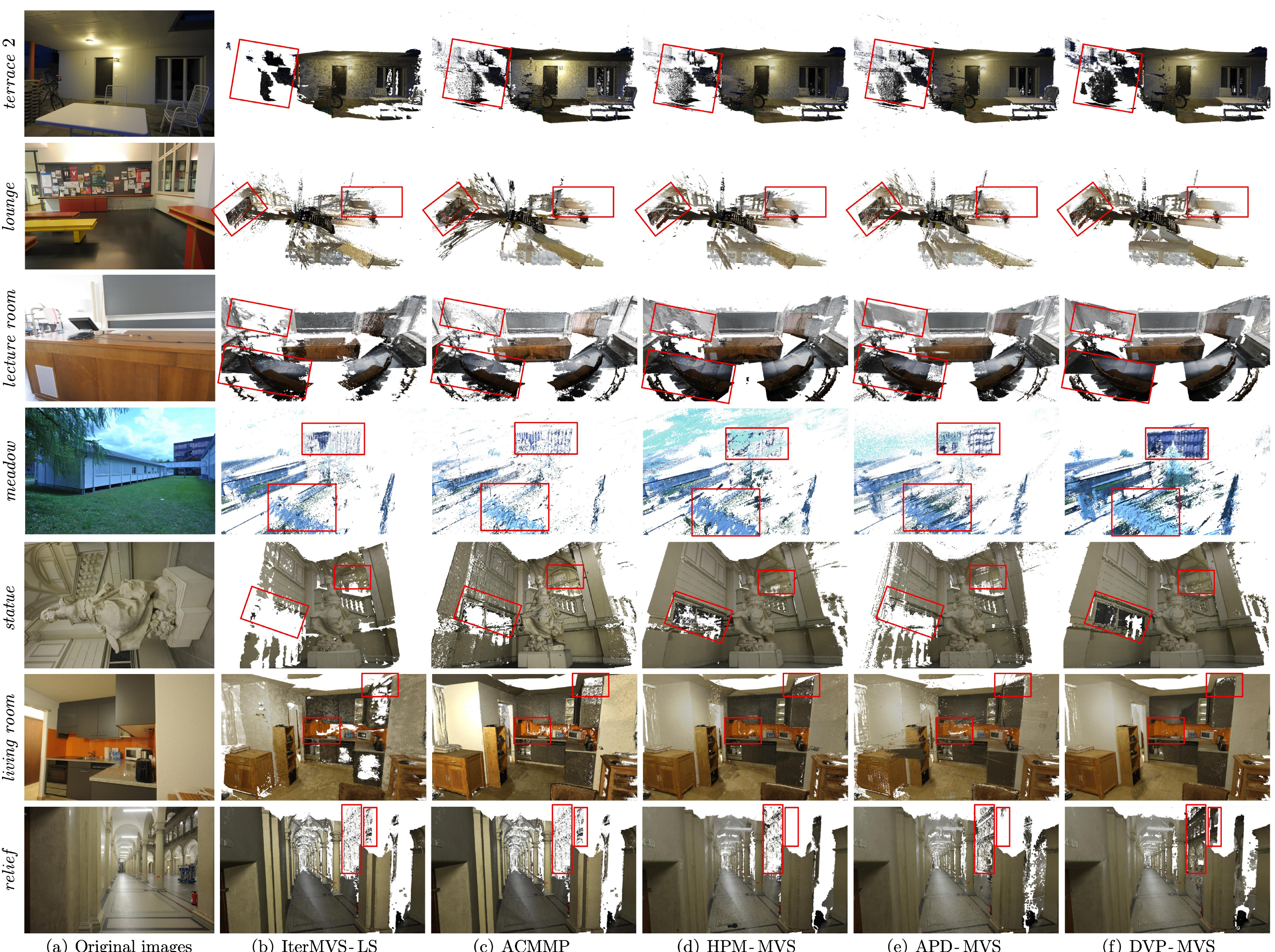}
    \caption{An illustration of the qualitative results on partial scenes of ETH3D datasets (\emph{terrace 2}, \emph{lounge}, \emph{lecture room}, \emph{meadow}, \emph{statue}, \emph{living room} and \emph{relief}). Our methods can effectively reconstruct large textureless areas as shown in red boxes.}
    \label{fig:eth3d_supp}
\end{figure*}

\begin{figure*}
    \centering
    \includegraphics[width=0.9\linewidth]{tnt_intermediate2.pdf}
    \caption{Reconstructed point clouds results on the intermediate set of Tanks \& Temples datasets.}
    \label{fig:tnt_intermediate}
\end{figure*}

\begin{figure*}
    \centering
    \includegraphics[width=0.9\linewidth]{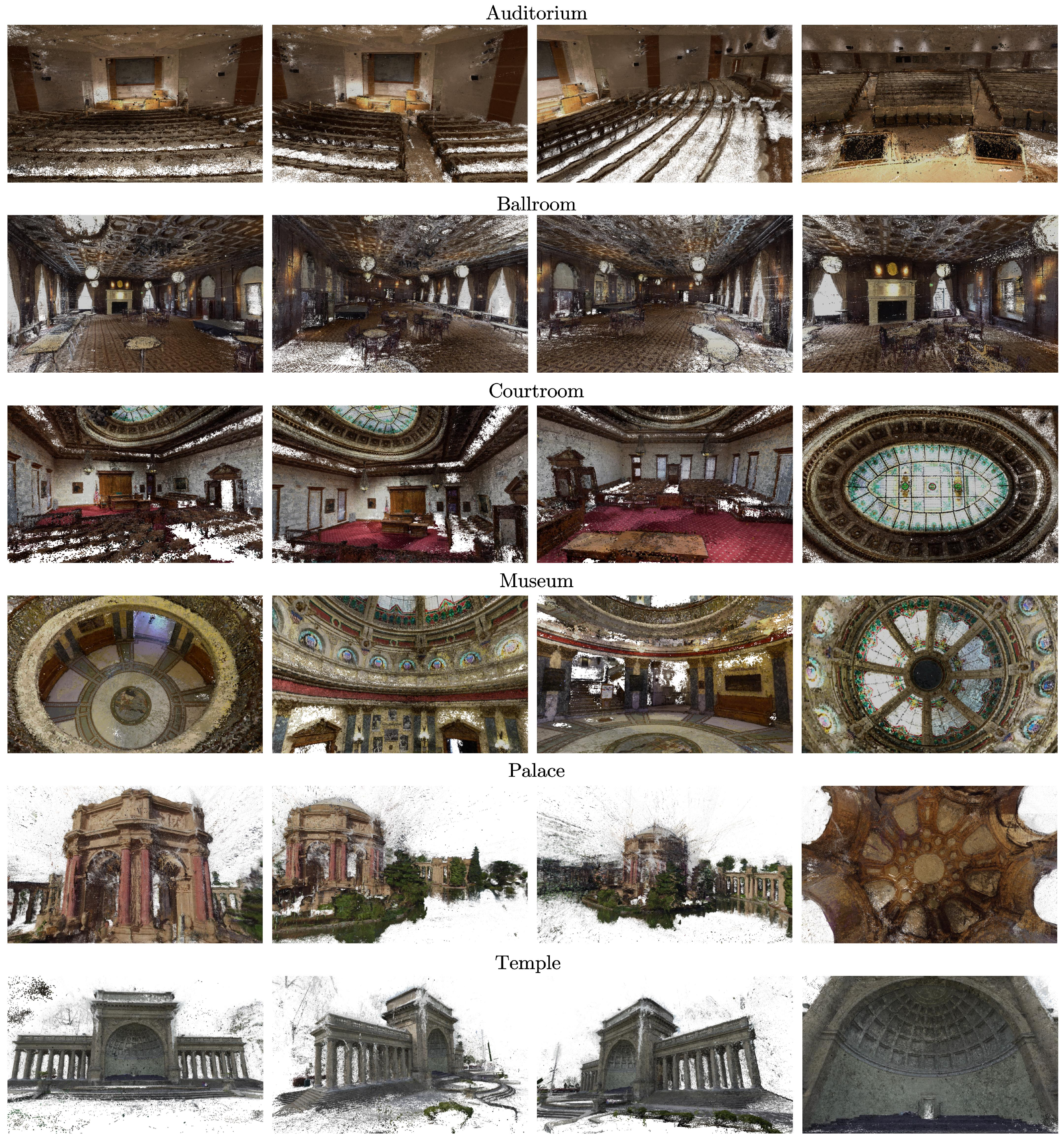}
    \caption{Reconstructed point clouds results on the advanced set of Tanks \& Temples datasets.}
    \label{fig:tnt_advanced}
\end{figure*}

\section{Expanded Related Work}
This section elaborates on the state-of-the-art in MVS, 3D scene representation, and the underlying geometric estimation techniques that form the foundation of our method.

\subsubsection{Recent Advances in Multi-View Stereo (MVS)}
The field of MVS is continuously advancing, with both traditional and learning-based methods pushing the envelope. While DVP-MVS enhances a traditional PatchMatch-based pipeline, a significant body of recent work has focused on novel neural architectures. For instance, researchers have explored Recurrent Regularization Transformers~\citep{rrt-mvs} and even State Space Models~\citep{mvsmamba} to better aggregate information across multiple views. Another prominent direction is leveraging monocular priors to guide the stereo matching process, which has proven effective in challenging scenarios~\citep{monomvsnet}. Our work, DVP-MVS, shares this philosophy of leveraging strong priors, though we focus on explicit geometric and visibility constraints rather than learned features. Building upon the principles established in DVP-MVS, our subsequent research has continued to refine this direction by incorporating dual-level precision edges and more accurate planarization techniques~\citep{chen2025dual}, as well as integrating segmentation-driven deformation with robust depth restoration and occlusion handling~\citep{yuan2025sed}. 

\subsubsection{Modern 3D Scene Representation and Rendering}
The output of MVS systems, typically a dense point cloud or depth maps, serves as the geometric foundation for modern 3D scene representation and rendering techniques. The advent of Neural Radiance Fields (NeRF) and, more recently, 3D Gaussian Splatting (3DGS) has revolutionized this space. The quality of the initial geometry is paramount for these methods, and a high-quality MVS reconstruction like that from DVP-MVS is invaluable. A significant amount of research is dedicated to improving 3DGS, focusing on aspects like lightweight compact representations for 4D generation~\citep{liu2025light4gs}, real-time dynamic scene rendering via spatio-temporal decoupling~\citep{li2025stdr}, and highly efficient video representation~\citep{liu2025d2gv}. Further advancements aim to enhance the structural integrity and boundary precision of Gaussian representations by leveraging topological priors~\citep{shen2025topology} or gradient guidance~\citep{li2024gradiseg}. Efforts are also underway to create controllable editing frameworks for 3DGS~\citep{yan20243dsceneeditor} and to unify appearance codes for complex driving scenes~\citep{wang2025unifying}. The development of comprehensive benchmarks for NeRF-based reconstruction further highlights the community's focus on evaluating and improving these 3D representations~\citep{yan2023nerfbk}.

\subsubsection{Advancements in Geometric Priors and Core Techniques}
DVP-MVS critically relies on high-quality priors, namely depth and edges. The field of self-supervised monocular depth estimation, is itself a vibrant area of research, with a series of works exploring curriculum contrastive learning for adverse weather~\citep{Wang_2024}, harnessing powerful diffusion models as priors~\citep{wang2025jasmine, Wang_2024_2}, and even reframing the task from an image editor's perspective~\citep{wang2025editor}. Similarly, the underlying task of stereo matching is continuously being improved with techniques like hybrid-supervised symmetric networks~\citep{Zhang2023EHSSAE}. Beyond depth, semantic segmentation provides an even richer prior for constraining patch deformation. Innovations in this area include dual-form complementary masking for domain adaptation~\citep{wang2025masktwins}, scaling up autoregressive pretraining for specialized domains like neuron segmentation~\citep{chen2025tokenunify}, and even using multi-agent reinforcement learning for self-supervised segmentation~\citep{chen2023self}. Furthermore, the fundamental problem of learning robust visual representations, crucial for matching, is being tackled through novel masking strategies in graph autoencoders~\citep{song2025spmgae, song2025equipping} and hierarchical cross-modal alignment for decoupled multimodal representations~\citep{qian2025decalign}.

\subsection{Broader Context and Future Directions}
This section situates 3D reconstruction within the wider landscape of artificial intelligence, connecting it to the paradigm shifts brought by large models and highlighting its pivotal role in various high-impact applications.

\subsubsection{The Era of Large Models: From Language to 3D Vision}
The transformative impact of Large Language Models (LLMs) and Vision-Language Models (VLMs) is reshaping all fields of AI. While DVP-MVS is a purely geometric method, the reasoning and world-knowledge capabilities of these large models present exciting future possibilities. A comprehensive body of work in LLMs actively explores improving multi-domain reasoning~\citep{bi2025reward}, enhancing confidence on edited facts~\citep{bi2024decoding}, controlling knowledge reliance~\citep{bi2025parameters}, and achieving better context-faithfulness~\citep{bi2024context}. Concurrently, methods are being developed to refine pre-training data at scale~\citep{bi2025refinex} and to create lightweight model transition techniques for LLM training~\citep{li2025wiscalightweightmodeltransition}. This intelligence is now being integrated into vision. VLMs are being challenged with complex spatial reasoning tasks~\citep{song2025siri} and endowed with explainability through diffusion-based chain-of-thought processes~\citep{lu2024explainable}. The paradigm of "prompting" is becoming central, with work on automatically optimizing prompts for classification~\citep{ProAPO} and leveraging retrieval-augmented preference optimization for better alignment~\citep{xing2025re}. This extends to zero-shot learning using multi-attribute document supervision~\citep{MADS} and visual-semantic decomposition~\citep{EmDepart}. For specific visual tasks, synergistic prompting is being used for human-object interaction (HOI) detection~\citep{luo2025synergistic}, which can be further improved with context-aware instructions~\citep{luoinstructhoi}. The generative and compositional power of these models is being applied to creative tasks like versatile advertising poster layout generation~\citep{anonymous2025anylayout}, and their ability to handle complex multi-modal inputs is being scaled up with architectures like the multi-modal diffusion mamba~\citep{lu2025end}. In the future, one could envision instructing MVS system with natural language, guided by the rich semantic understanding.

\subsubsection{Advanced Learning Paradigms and Applications}
Beyond large models, other advanced learning paradigms hold potential for future MVS systems. For instance, diffusion-driven data generation is being used to enhance object detection, with specific applications in image generation~\citep{tang2024crs} and broader data augmentation~\citep{tang2025aerogen}. Graph learning continues to be a powerful tool for tasks ranging from anti-sparse downscaling~\citep{fan2025multi} to ensuring the resilience of graph prompt tuning against attacks~\citep{song2025gpromptshield}. The robustness and security of models are also gaining attention, exemplified by the development of watermarking frameworks resistant to extreme cropping~\citep{sunultra} and non-differentiable distortions~\citep{sun2025end}. This focus on robustness is critical for deploying MVS in real-world applications. Connecting 3D vision to downstream tasks is a major driver of research. The field of composed image and video retrieval, for example, seeks to understand fine-grained visual modifications, requiring a suite of methods including hierarchical disambiguation networks~\citep{HUD}, segmentation-based focus shift revision~\citep{OFFSET}, explicit semantic parsing~\citep{FineCIR}, and complementarity-guided disentanglement~\citep{PAIR}. These tasks rely on a deep understanding of visual entities and relations~\citep{ENCODER}, which a precise 3D model can greatly facilitate. Similarly, understanding human-object interactions by discovering syntactic clues~\citep{luo2024discovering} or recognizing actions with the help of external tools~\citep{yuan2025video} are higher-level reasoning tasks that are grounded in the 3D world reconstructed by MVS.

\subsubsection{Cross-Disciplinary Impact of 3D Vision}
High-fidelity 3D reconstruction is most evident in its wide-ranging impact across various scientific and industrial domains.

\paragraph{Autonomous Driving and Robotics.} This is a primary application area. MVS is fundamental for building the high-definition (HD) maps that autonomous vehicles rely on. Current research focuses on online HD map construction~\citep{zhang2025mapexpert} and developing end-to-end driving models that exploit adversarial transfer~\citep{zhangdrive} or use distinct experts with implicit interactions~\citep{zhangaddi}. World models with self-supervised 3D labels are being developed for simulation and planning~\citep{yan2025renderworld}, and vision-language agents are being trained with reasoning and self-reflection for driving tasks~\citep{yuan2025autodrive}, which are part of a broader survey on pure Vision Language Action (VLA) models~\citep{zhang2025pure}. Trajectory prediction using shared 3D queries is also a key component that benefits from 3D scene understanding~\citep{song2023xvtp3d}. In robotics, physical autoregressive models are enabling manipulation without pre-training~\citep{song2025physical}. A cloud framework with autonomous driving path planning can leverage edge-cloud resources for complex computations~\citep{10858860}.

\paragraph{Remote Sensing.} MVS and related 3D techniques are crucial for interpreting satellite and aerial imagery. Research in this area includes developing comprehensive surveys of spatiotemporal vision-language models for remote sensing~\citep{liu2025RSTVLM_survey} and building interactive agents for change interpretation~\citep{liu2024changeAgent}. Meanwhile, a significant effort is dedicated to data analysis, where advanced multi-view graph clustering techniques are being specifically tailored for remote sensing data, with innovations in dual relation optimization~\citep{MDRO}, structure-adaptive mechanisms~\citep{SAMVGC}, and contrastive learning with long-short range information mining~\citep{SEC-LSRM}.

\paragraph{Medical Image Analysis.} This is another high-impact domain. 3D reconstruction and analysis are vital for interpreting volumetric scans. Large models are being adapted for medical applications, such as using evolutionary algorithms for prompt optimization in diagnostics~\citep{11205280} and generative text-guided 3D pretraining for segmentation~\citep{chen2025gtgm}. A significant challenge is handling imbalanced data and noisy labels, which is being addressed by a suite of methods including curriculum learning frameworks~\citep{han2025climd, qian2025dyncim} and adaptive label correction techniques~\citep{qian2025adaptive}. Researchers are also exploring region uncertainty estimation~\citep{han2025region} and leveraging the frequency domain for segmentation tasks~\citep{han2025frequency}. In pathology, sparse hierarchical transformers are used for survival analysis on whole slide images~\citep{10226279}, and multi-scale foundation models are being fused for comprehensive analysis~\citep{yang2025fusionmultiscaleheterogeneouspathology}. Furthermore, a series of specialized denoising techniques based on noise modeling, sparsity constraints, and simulation-aware pretraining have been developed to improve the quality of Cryo-Electron Tomography and Microscopy data~\citep{Yang_2021_ICCV, nmsg, n2tc, cmb20240513}. Even in more traditional industrial contexts, methods for classifying steel surface defects based on salient local features are being developed~\citep{hao2024ssdc}.

Finally, the ultimate goal is to create systems that can seamlessly collaborate with humans. World cognition agents enabling adaptive human-AI symbiosis in Industry 4.0 represent a forward-looking vision where the 3D understanding provided by systems like DVP-MVS becomes a cornerstone of intelligent interaction~\citep{liu2025mr}.

\subsection{Conclusion}
In this expanded review, we have situated DVP-MVS within a comprehensive network of contemporary research. We began by deepening the discussion on its core domain, MVS and 3D representation, and then broadened the horizon to encompass the paradigm-shifting influence of large models and the diverse applications of 3D vision. This exploration underscores a key insight: while DVP-MVS stands as a testament to the power of robust, geometrically-grounded principles, its future impact will be magnified through synergistic integration with the semantic reasoning of large models and by serving as a foundational technology for high-stakes domains like autonomous driving and medical imaging. The exciting path forward lies at the intersection of these fields, where precise geometry and deep semantic understanding converge to create truly intelligent systems.

\bibliography{aaai25}

\begin{thebibliography}{137}
\providecommand{\natexlab}[1]{#1}

\bibitem[{Anonymous(2025)}]{anonymous2025anylayout}
Anonymous. 2025.
\newblock AnyLayout: Versatile Advertising Poster Layout Generation with {MLLM}s.
\newblock In \emph{Submitted to The Fourteenth International Conference on Learning Representations}.
\newblock Under review.

\bibitem[{Barnes et~al.(2009)Barnes, Shechtman, Finkelstein, and Goldman}]{PM}
Barnes, C.; Shechtman, E.; Finkelstein, A.; and Goldman, D.~B. 2009.
\newblock PatchMatch: A randomized correspondence algorithm for structural image editing.
\newblock \emph{ACM Trans. Graph.}, 28(3): 24.

\bibitem[{Bi et~al.(2024{\natexlab{a}})Bi, Huang, Wang, Yang, Zhang, Huang, Mei, Fang, Li, Wei et~al.}]{bi2024context}
Bi, B.; Huang, S.; Wang, Y.; Yang, T.; Zhang, Z.; Huang, H.; Mei, L.; Fang, J.; Li, Z.; Wei, F.; et~al. 2024{\natexlab{a}}.
\newblock Context-DPO: Aligning Language Models for Context-Faithfulness.
\newblock \emph{ACL 2025}.

\bibitem[{Bi et~al.(2024{\natexlab{b}})Bi, Liu, Mei, Wang, Ji, and Cheng}]{bi2024decoding}
Bi, B.; Liu, S.; Mei, L.; Wang, Y.; Ji, P.; and Cheng, X. 2024{\natexlab{b}}.
\newblock Decoding by Contrasting Knowledge: Enhancing LLMs' Confidence on Edited Facts.
\newblock \emph{ACL 2025}.

\bibitem[{Bi et~al.(2025{\natexlab{a}})Bi, Liu, Ren, Liu, Lin, Wang, Mei, Fang, Guo, and Cheng}]{bi2025refinex}
Bi, B.; Liu, S.; Ren, X.; Liu, D.; Lin, J.; Wang, Y.; Mei, L.; Fang, J.; Guo, J.; and Cheng, X. 2025{\natexlab{a}}.
\newblock RefineX: Learning to Refine Pre-training Data at Scale from Expert-Guided Programs.
\newblock \emph{arXiv preprint arXiv:2507.03253}.

\bibitem[{Bi et~al.(2025{\natexlab{b}})Bi, Liu, Wang, Tong, Mei, Ge, Xu, Guo, and Cheng}]{bi2025reward}
Bi, B.; Liu, S.; Wang, Y.; Tong, S.; Mei, L.; Ge, Y.; Xu, Y.; Guo, J.; and Cheng, X. 2025{\natexlab{b}}.
\newblock Reward and Guidance through Rubrics: Promoting Exploration to Improve Multi-Domain Reasoning.
\newblock \emph{arXiv preprint arXiv:2511.12344}.

\bibitem[{Bi et~al.(2025{\natexlab{c}})Bi, Liu, Wang, Xu, Fang, Mei, and Cheng}]{bi2025parameters}
Bi, B.; Liu, S.; Wang, Y.; Xu, Y.; Fang, J.; Mei, L.; and Cheng, X. 2025{\natexlab{c}}.
\newblock Parameters vs. context: Fine-grained control of knowledge reliance in language models.
\newblock \emph{arXiv preprint arXiv:2503.15888}.

\bibitem[{Bleyer, Rhemann, and Rother(2011)}]{PMS}
Bleyer, M.; Rhemann, C.; and Rother, C. 2011.
\newblock PatchMatch Stereo - Stereo Matching with Slanted Support Windows.
\newblock In Hoey, J.; McKenna, S.~J.; and Trucco, E., eds., \emph{British Mach. Vis. Conf. (BMVC)}, 1--11.

\bibitem[{Chen et~al.(2025{\natexlab{a}})Chen, Yuan, Mao, and Wang}]{chen2025dual}
Chen, K.; Yuan, Z.; Mao, T.; and Wang, Z. 2025{\natexlab{a}}.
\newblock Dual-level precision edges guided multi-view stereo with accurate planarization.
\newblock In \emph{Proceedings of the AAAI Conference on Artificial Intelligence}, volume~39, 2105--2113.

\bibitem[{Chen et~al.(2025{\natexlab{b}})Chen, He, Yang, Zhang, Yuan, Khan, Baili, and Yee}]{11205280}
Chen, Y.; He, Y.; Yang, J.; Zhang, D.; Yuan, Z.; Khan, M.~A.; Baili, J.; and Yee, P.~L. 2025{\natexlab{b}}.
\newblock EMPOWER: Evolutionary Medical Prompt Optimization With Reinforcement Learning.
\newblock \emph{IEEE Journal of Biomedical and Health Informatics}, 1--10.

\bibitem[{Chen et~al.(2024{\natexlab{a}})Chen, Huang, Liu, Deng, Chen, and Xiong}]{chen2024learning}
Chen, Y.; Huang, W.; Liu, X.; Deng, S.; Chen, Q.; and Xiong, Z. 2024{\natexlab{a}}.
\newblock Learning multiscale consistency for self-supervised electron microscopy instance segmentation.
\newblock In \emph{ICASSP}, 1566--1570. IEEE.

\bibitem[{Chen et~al.(2023)Chen, Huang, Zhou, Chen, and Xiong}]{chen2023self}
Chen, Y.; Huang, W.; Zhou, S.; Chen, Q.; and Xiong, Z. 2023.
\newblock Self-supervised neuron segmentation with multi-agent reinforcement learning.
\newblock In \emph{International Joint Conference on Artificial Intelligence (IJCAI)}.

\bibitem[{Chen et~al.(2025{\natexlab{c}})Chen, Liu, Huang, Liu, Shi, Cheng, Arcucci, and Xiong}]{chen2025gtgm}
Chen, Y.; Liu, C.; Huang, W.; Liu, X.; Shi, H.; Cheng, S.; Arcucci, R.; and Xiong, Z. 2025{\natexlab{c}}.
\newblock GTGM: Generative Text-Guided 3D Vision-Language Pretraining for Medical Image Segmentation.
\newblock In \emph{IEEE International Conference on Computer Vision Workshop on Vision Language Models for 3D Understanding (ICCV Workshop VLM3D)}, 6715--6724.

\bibitem[{Chen et~al.(2024{\natexlab{b}})Chen, Liu, Liu, Arcucci, and Xiong}]{chen2024bimcv}
Chen, Y.; Liu, C.; Liu, X.; Arcucci, R.; and Xiong, Z. 2024{\natexlab{b}}.
\newblock Bimcv-r: A landmark dataset for 3d ct text-image retrieval.
\newblock In \emph{MICCAI}, 124--134. Springer.

\bibitem[{Chen et~al.(2024{\natexlab{c}})Chen, Shi, Liu, Shi, Zhang, Liu, Xiong, and Wu}]{chen2024tokenunify}
Chen, Y.; Shi, H.; Liu, X.; Shi, T.; Zhang, R.; Liu, D.; Xiong, Z.; and Wu, F. 2024{\natexlab{c}}.
\newblock TokenUnify: Scalable Autoregressive Visual Pre-training with Mixture Token Prediction.
\newblock \emph{arXiv preprint arXiv:2405.16847}.

\bibitem[{Chen et~al.(2025{\natexlab{d}})Chen, Shi, Liu, Shi, Zhang, Liu, Xiong, and Wu}]{chen2025tokenunify}
Chen, Y.; Shi, H.; Liu, X.; Shi, T.; Zhang, R.; Liu, D.; Xiong, Z.; and Wu, F. 2025{\natexlab{d}}.
\newblock TokenUnify: Scaling Up Autoregressive Pretraining for Neuron Segmentation.
\newblock In \emph{IEEE International Conference on Computer Vision (ICCV)}, 13604--13613.

\bibitem[{Chen et~al.(2025{\natexlab{e}})Chen, Hu, Li, Fu, Song, and Nie}]{OFFSET}
Chen, Z.; Hu, Y.; Li, Z.; Fu, Z.; Song, X.; and Nie, L. 2025{\natexlab{e}}.
\newblock OFFSET: Segmentation-based Focus Shift Revision for Composed Image Retrieval.
\newblock In \emph{Proceedings of the ACM International Conference on Multimedia}, 6113–6122.

\bibitem[{Chen et~al.(2025{\natexlab{f}})Chen, Hu, Li, Fu, Wen, and Guan}]{HUD}
Chen, Z.; Hu, Y.; Li, Z.; Fu, Z.; Wen, H.; and Guan, W. 2025{\natexlab{f}}.
\newblock HUD: Hierarchical Uncertainty-Aware Disambiguation Network for Composed Video Retrieval.
\newblock In \emph{Proceedings of the ACM International Conference on Multimedia}, 6143–6152.

\bibitem[{Cheng et~al.(2021)Cheng, Li, Zhang, Xia, and Zhang}]{Cheng2}
Cheng, D.; Li, S.; Zhang, H.; Xia, F.; and Zhang, Y. 2021.
\newblock Why Dataset Properties Bound the Scalability of Parallel Machine Learning Training Algorithms.
\newblock \emph{IEEE Transactions on Parallel and Distributed Systems}, 32(7): 1702--1712.

\bibitem[{Cheng, Li, and Zhang(2020)}]{Cheng1}
Cheng, D.; Li, S.; and Zhang, Y. 2020.
\newblock {{WP-SGD}}: {{Weighted}} Parallel {{SGD}} for Distributed Unbalanced-Workload Training System.
\newblock \emph{Journal of Parallel and Distributed Computing}, 145: 202--216.

\bibitem[{Cheng, Li, and Zhang(2023)}]{Cheng3}
Cheng, D.; Li, S.; and Zhang, Y. 2023.
\newblock Asynch-{{SGBDT}}: {{Train Stochastic Gradient Boosting Decision Trees}} in an {{Asynchronous Parallel Manner}}.
\newblock In \emph{2023 {{IEEE International Parallel}} and {{Distributed Processing Symposium}} ({{IPDPS}})}, 256--267. IEEE.

\bibitem[{Cheng and Sun(2024)}]{SUN1}
Cheng, S.; and Sun, H. 2024.
\newblock SPT: Sequence Prompt Transformer for Interactive Image Segmentation.
\newblock arXiv:2412.10224.

\bibitem[{Ding et~al.(2023)Ding, Fan, Yang, and Xia}]{DING1}
Ding, B.; Fan, Z.; Yang, S.; and Xia, S. 2023.
\newblock MyPortrait: Morphable Prior-Guided Personalized Portrait Generation.
\newblock \emph{arXiv preprint arXiv:2312.02703}.

\bibitem[{Ding et~al.(2024)Ding, Fan, Zhao, and Xia}]{DING2}
Ding, B.; Fan, Z.; Zhao, Z.; and Xia, S. 2024.
\newblock Mining collaborative spatio-temporal clues for face forgery detection.
\newblock \emph{Multimedia Tools and Applications}, 83(9): 27901--27920.

\bibitem[{Fan et~al.(2025)Fan, Yu, Barclay, Appling, Sun, Xie, and Jia}]{fan2025multi}
Fan, Y.; Yu, R.; Barclay, J.~R.; Appling, A.~P.; Sun, Y.; Xie, Y.; and Jia, X. 2025.
\newblock Multi-Scale Graph Learning for Anti-Sparse Downscaling.
\newblock In \emph{Proceedings of the AAAI Conference on Artificial Intelligence}, volume~39, 27969--27977.

\bibitem[{Fu et~al.(2025)Fu, Li, Chen, Wang, Song, Hu, and Nie}]{PAIR}
Fu, Z.; Li, Z.; Chen, Z.; Wang, C.; Song, X.; Hu, Y.; and Nie, L. 2025.
\newblock PAIR: Complementarity-guided Disentanglement for Composed Image Retrieval.
\newblock In \emph{Proceedings of the IEEE International Conference on Acoustics, Speech and Signal Processing}, 1--5. IEEE.

\bibitem[{Galliani, Lasinger, and Schindler(2015)}]{Galliani}
Galliani, S.; Lasinger, K.; and Schindler, K. 2015.
\newblock Massively Parallel Multiview Stereopsis by Surface Normal Diffusion.
\newblock In \emph{Proc. IEEE/CVF Int. Conf. Comput. Vis. (ICCV)}.

\bibitem[{Gan et~al.(2024)Gan, Liu, Xu, Mo, and Yokoya}]{Gan2024a}
Gan, W.; Liu, F.; Xu, H.; Mo, N.; and Yokoya, N. 2024.
\newblock {{GaussianOcc}}: {{Fully Self-supervised}} and {{Efficient 3D Occupancy Estimation}} with {{Gaussian Splatting}}.
\newblock arXiv:2408.11447.

\bibitem[{Gan et~al.(2021)Gan, Wong, Yu, Zhao, and Vong}]{Gan2021}
Gan, W.; Wong, P.~K.; Yu, G.; Zhao, R.; and Vong, C.~M. 2021.
\newblock Light-Weight Network for Real-Time Adaptive Stereo Depth Estimation.
\newblock \emph{Neurocomputing}, 441: 118--127.

\bibitem[{Guan et~al.(2025{\natexlab{a}})Guan, Li, Wang, Tu, Li, Zhu, Liu, and Chen}]{MDRO}
Guan, R.; Li, J.; Wang, S.; Tu, W.; Li, M.; Zhu, E.; Liu, X.; and Chen, P. 2025{\natexlab{a}}.
\newblock Multi-view Graph Clustering with Dual Relation Optimization for Remote Sensing Data.
\newblock In \emph{Proceedings of the 33rd ACM International Conference on Multimedia}, 7346–7355.

\bibitem[{Guan et~al.(2024{\natexlab{a}})Guan, Li, Tu, Wang, Liu, Li, Tang, and Feng}]{GUAN1}
Guan, R.; Li, Z.; Tu, W.; Wang, J.; Liu, Y.; Li, X.; Tang, C.; and Feng, R. 2024{\natexlab{a}}.
\newblock Contrastive Multiview Subspace Clustering of Hyperspectral Images Based on Graph Convolutional Networks.
\newblock \emph{IEEE Transactions on Geoscience and Remote Sensing}, 62: 1--14.

\bibitem[{Guan et~al.(2025{\natexlab{b}})Guan, Liu, Tu, Tang, Luo, and Liu}]{SEC-LSRM}
Guan, R.; Liu, T.; Tu, W.; Tang, C.; Luo, W.; and Liu, X. 2025{\natexlab{b}}.
\newblock Sampling Enhanced Contrastive Multi-View Remote Sensing Data Clustering with Long-Short Range Information Mining.
\newblock \emph{IEEE Transactions on Knowledge and Data Engineering}, 1--15.

\bibitem[{Guan et~al.(2024{\natexlab{b}})Guan, Tu, Li, Yu, Hu, Chen, Tang, Yuan, and Liu}]{GUAN2}
Guan, R.; Tu, W.; Li, Z.; Yu, H.; Hu, D.; Chen, Y.; Tang, C.; Yuan, Q.; and Liu, X. 2024{\natexlab{b}}.
\newblock Spatial-Spectral Graph Contrastive Clustering with Hard Sample Mining for Hyperspectral Images.
\newblock \emph{IEEE Transactions on Geoscience and Remote Sensing}, 1--16.

\bibitem[{Guan et~al.(2025{\natexlab{c}})Guan, Tu, Wang, Liu, Hu, Tang, Feng, Li, Xiao, and Liu}]{SAMVGC}
Guan, R.; Tu, W.; Wang, S.; Liu, J.; Hu, D.; Tang, C.; Feng, Y.; Li, J.; Xiao, B.; and Liu, X. 2025{\natexlab{c}}.
\newblock Structure-Adaptive Multi-View Graph Clustering for Remote Sensing Data.
\newblock In \emph{Proceedings of the AAAI Conference on Artificial Intelligence}, volume~39, 16933--16941.

\bibitem[{Han et~al.(2025{\natexlab{a}})Han, Lyu, Ma, Qian, Ma, Pang, Chen, and Liu}]{han2025climd}
Han, K.; Lyu, C.; Ma, L.; Qian, C.; Ma, S.; Pang, Z.; Chen, J.; and Liu, Z. 2025{\natexlab{a}}.
\newblock Climd: A curriculum learning framework for imbalanced multimodal diagnosis.
\newblock In \emph{International Conference on Medical Image Computing and Computer-Assisted Intervention}, 65--74. Springer.

\bibitem[{Han et~al.(2025{\natexlab{b}})Han, Ma, Qian, Chen, Lyu, Song, and Liu}]{han2025frequency}
Han, K.; Ma, S.; Qian, C.; Chen, J.; Lyu, C.; Song, Y.; and Liu, Z. 2025{\natexlab{b}}.
\newblock Frequency Domain Unlocks New Perspectives for Abdominal Medical Image Segmentation.
\newblock \emph{arXiv preprint arXiv:2510.11005}.

\bibitem[{Han et~al.(2025{\natexlab{c}})Han, Wang, Chen, Qian, Lyu, Ma, Qiu, Sheng, Huang, and Liu}]{han2025region}
Han, K.; Wang, S.; Chen, J.; Qian, C.; Lyu, C.; Ma, S.; Qiu, C.; Sheng, V.~S.; Huang, Q.; and Liu, Z. 2025{\natexlab{c}}.
\newblock Region uncertainty estimation for medical image segmentation with noisy labels.
\newblock \emph{IEEE Transactions on Medical Imaging}.

\bibitem[{Hao et~al.(2024)Hao, Gan, Liu, Chen, Shen, Qian, and Liu}]{hao2024ssdc}
Hao, Q.; Gan, Q.; Liu, Z.; Chen, J.; Shen, Q.; Qian, C.; and Liu, Y. 2024.
\newblock SSDC-Net: An Effective Classification Method of Steel Surface Defects Based on Salient Local Features.
\newblock In \emph{International Conference on Intelligent Computing}, 490--503. Springer.

\bibitem[{Jiang et~al.(2025{\natexlab{a}})Jiang, Liu, Liu, Yu, Wang, Chen, and Ma}]{mvsmamba}
Jiang, J.; Liu, Q.; Liu, H.; Yu, H.; Wang, L.; Chen, J.; and Ma, H. 2025{\natexlab{a}}.
\newblock MVSMamba: Multi-View Stereo with State Space Model.
\newblock \emph{arXiv preprint arXiv:2511.01315}.

\bibitem[{Jiang et~al.(2025{\natexlab{b}})Jiang, Liu, Yu, Liu, Wang, Chen, and Ma}]{monomvsnet}
Jiang, J.; Liu, Q.; Yu, H.; Liu, H.; Wang, L.; Chen, J.; and Ma, H. 2025{\natexlab{b}}.
\newblock MonoMVSNet: Monocular Priors Guided Multi-View Stereo Network.
\newblock In \emph{Proceedings of the IEEE/CVF International Conference on Computer Vision (ICCV)}, 27806--27816.

\bibitem[{Jiang et~al.(2025{\natexlab{c}})Jiang, Wang, Yu, Hu, Chen, and Ma}]{rrt-mvs}
Jiang, J.; Wang, L.; Yu, H.; Hu, T.; Chen, J.; and Ma, H. 2025{\natexlab{c}}.
\newblock RRT-MVS: Recurrent Regularization Transformer for Multi-View Stereo.
\newblock In \emph{Proceedings of the AAAI Conference on Artificial Intelligence}, volume~39, 3994--4002.

\bibitem[{Knapitsch et~al.(2017)Knapitsch, Park, Zhou, and Koltun}]{TNT}
Knapitsch, A.; Park, J.; Zhou, Q.-Y.; and Koltun, V. 2017.
\newblock Tanks and Temples: {{Benchmarking}} Large-Scale Scene Reconstruction.
\newblock \emph{ACM Transactions on Graphics (ToG)}, 36(4): 1--13.

\bibitem[{Li et~al.(2022{\natexlab{a}})Li, Zhang, Wan, Yang, Li, Li, Han, Zhu, and Zhang}]{Li2022b}
Li, H.; Zhang, H.; Wan, X.; Yang, Z.; Li, C.; Li, J.; Han, R.; Zhu, P.; and Zhang, F. 2022{\natexlab{a}}.
\newblock Noise-{{Transfer2Clean}}: Denoising Cryo-{{EM}} Images Based on Noise Modeling and Transfer.
\newblock \emph{Bioinformatics}, 38(7): 2022--2029.

\bibitem[{Li et~al.(2022{\natexlab{b}})Li, Zhang, Wan, Yang, Li, Li, Han, Zhu, and Zhang}]{n2tc}
Li, H.; Zhang, H.; Wan, X.; Yang, Z.; Li, C.; Li, J.; Han, R.; Zhu, P.; and Zhang, F. 2022{\natexlab{b}}.
\newblock {Noise-Transfer2Clean: denoising cryo-EM images based on noise modeling and transfer}.
\newblock \emph{Bioinformatics}, 38(7): 2022--2029.

\bibitem[{Li et~al.(2025{\natexlab{a}})Li, Tan, Yang, Sun, Huo, Qin, Sun, Xie, Cai, Zhang, He, Tan, Jia, and Zhao}]{li2025wiscalightweightmodeltransition}
Li, J.; Tan, J.; Yang, Z.; Sun, P.; Huo, F.; Qin, J.; Sun, Y.; Xie, Y.; Cai, X.; Zhang, X.; He, M.; Tan, G.; Jia, W.; and Zhao, T. 2025{\natexlab{a}}.
\newblock WISCA: A Lightweight Model Transition Method to Improve LLM Training via Weight Scaling.

\bibitem[{Li et~al.(2024{\natexlab{a}})Li, Zhang, Lei, Zhao, Li, Kamnoedboon, Dong, and Li}]{XIMAGENET}
Li, Q.; Zhang, D.; Lei, S.; Zhao, X.; Li, W.; Kamnoedboon, P.; Dong, J.; and Li, S. 2024{\natexlab{a}}.
\newblock {{XIMAGENET-12}}: {{An Explainable Visual Benchmark Dataset}} for {{Model Robustness Evaluation}}.
\newblock In \emph{Synthetic {{Data}} for {{Computer Vision Workshop}}@ {{CVPR}} 2024}.

\bibitem[{Li et~al.(2025{\natexlab{b}})Li, Chen, Wen, Fu, Hu, and Guan}]{ENCODER}
Li, Z.; Chen, Z.; Wen, H.; Fu, Z.; Hu, Y.; and Guan, W. 2025{\natexlab{b}}.
\newblock Encoder: Entity mining and modification relation binding for composed image retrieval.
\newblock In \emph{Proceedings of the AAAI Conference on Artificial Intelligence}, volume~39, 5101--5109.

\bibitem[{Li et~al.(2025{\natexlab{c}})Li, Fu, Hu, Chen, Wen, and Nie}]{FineCIR}
Li, Z.; Fu, Z.; Hu, Y.; Chen, Z.; Wen, H.; and Nie, L. 2025{\natexlab{c}}.
\newblock FineCIR: Explicit Parsing of Fine-Grained Modification Semantics for Composed Image Retrieval.
\newblock \emph{https://arxiv.org/abs/2503.21309}.

\bibitem[{Li et~al.(2024{\natexlab{b}})Li, Han, Cai, Jiang, Bi, Gao, Zhao, and Wang}]{li2024gradiseg}
Li, Z.; Han, W.; Cai, Y.; Jiang, H.; Bi, B.; Gao, S.; Zhao, H.; and Wang, Z. 2024{\natexlab{b}}.
\newblock Gradiseg: Gradient-guided gaussian segmentation with enhanced 3d boundary precision.
\newblock \emph{arXiv preprint arXiv:2412.00392}.

\bibitem[{Li et~al.(2025{\natexlab{d}})Li, Jiang, Cai, Chen, Bi, Gao, Zhao, Wang, Mao, and Wang}]{li2025stdr}
Li, Z.; Jiang, H.; Cai, Y.; Chen, J.; Bi, B.; Gao, S.; Zhao, H.; Wang, Y.; Mao, T.; and Wang, Z. 2025{\natexlab{d}}.
\newblock STDR: Spatio-Temporal Decoupling for Real-Time Dynamic Scene Rendering.
\newblock \emph{arXiv preprint arXiv:2505.22400}.

\bibitem[{Liao et~al.(2019)Liao, Fu, Yan, and Xiao}]{Pyramid}
Liao, J.; Fu, Y.; Yan, Q.; and Xiao, C. 2019.
\newblock Pyramid {{Multi}}-{{View Stereo}} with {{Local Consistency}}.
\newblock \emph{Computer Graphics Forum}, 38(7): 335--346.

\bibitem[{Liu et~al.(2024)Liu, Chen, Zhang, Qi, Zou, and Shi}]{liu2024changeAgent}
Liu, C.; Chen, K.; Zhang, H.; Qi, Z.; Zou, Z.; and Shi, Z. 2024.
\newblock Change-Agent: Toward Interactive Comprehensive Remote Sensing Change Interpretation and Analysis.
\newblock \emph{IEEE Transactions on Geoscience and Remote Sensing}, 62: 1--16.

\bibitem[{Liu et~al.(2025{\natexlab{a}})Liu, Yuan, Wang, Yin, Luo, He, and Liang}]{liu2025mr}
Liu, C.; Yuan, Z.; Wang, Y.; Yin, Y.; Luo, W.; He, Z.; and Liang, X. 2025{\natexlab{a}}.
\newblock MR-IntelliAssist: A World Cognition Agent Enabling Adaptive Human-AI Symbiosis in Industry 4.0.
\newblock In \emph{International Conference on Human-Computer Interaction}, 163--177. Springer Nature Switzerland Cham.

\bibitem[{Liu et~al.(2025{\natexlab{b}})Liu, Zhang, Chen, Wang, Zou, and Shi}]{liu2025RSTVLM_survey}
Liu, C.; Zhang, J.; Chen, K.; Wang, M.; Zou, Z.; and Shi, Z. 2025{\natexlab{b}}.
\newblock Remote Sensing Spatiotemporal Vision–Language Models: A comprehensive survey.
\newblock \emph{IEEE Geoscience and Remote Sensing Magazine}, 2--42.

\bibitem[{Liu et~al.(2025{\natexlab{c}})Liu, Yang, Huang, Huang, Yuan, Li, and Xu}]{liu2025light4gs}
Liu, M.; Yang, Q.; Huang, H.; Huang, W.; Yuan, Z.; Li, Z.; and Xu, Y. 2025{\natexlab{c}}.
\newblock Light4gs: Lightweight compact 4d gaussian splatting generation via context model.
\newblock \emph{arXiv preprint arXiv:2503.13948}.

\bibitem[{Liu et~al.(2025{\natexlab{d}})Liu, Yang, Zhao, Huang, Yang, Li, and Xu}]{liu2025d2gv}
Liu, M.; Yang, Q.; Zhao, M.; Huang, H.; Yang, L.; Li, Z.; and Xu, Y. 2025{\natexlab{d}}.
\newblock D2gv: Deformable 2d gaussian splatting for video representation in 400fps.
\newblock \emph{arXiv preprint arXiv:2503.05600}.

\bibitem[{Lu et~al.(2025)Lu, Lu, Dong, and Luo}]{lu2025end}
Lu, C.; Lu, Q.; Dong, M.; and Luo, J. 2025.
\newblock End-to-End Multi-Modal Diffusion Mamba.
\newblock In \emph{Proceedings of the IEEE/CVF International Conference on Computer Vision}, 20529--20540.

\bibitem[{Lu, Lu, and Luo(2024)}]{lu2024explainable}
Lu, C.; Lu, Q.; and Luo, J. 2024.
\newblock An Explainable Vision Question Answer Model via Diffusion Chain-of-Thought.
\newblock In \emph{European Conference on Computer Vision}, 146--162. Springer.

\bibitem[{Luo et~al.(2024)Luo, Ren, Jiang, Chen, Wang, Han, and Liu}]{luo2024discovering}
Luo, J.; Ren, W.; Jiang, W.; Chen, X.; Wang, Q.; Han, Z.; and Liu, H. 2024.
\newblock Discovering syntactic interaction clues for human-object interaction detection.
\newblock In \emph{Proceedings of the IEEE/CVF Conference on Computer Vision and Pattern Recognition}, 28212--28222.

\bibitem[{Luo et~al.(2025{\natexlab{a}})Luo, Ren, Wang, Chen, Fan, Han, and Liu}]{luo2025synergistic}
Luo, J.; Ren, W.; Wang, Z.; Chen, X.; Fan, H.; Han, Z.; and Liu, H. 2025{\natexlab{a}}.
\newblock Synergistic Prompting Learning for Human-Object Interaction Detection.
\newblock \emph{IEEE Transactions on Image Processing}.

\bibitem[{Luo et~al.(2025{\natexlab{b}})Luo, Ren, Zheng, Zhang, Yuan, Wang, Lu, and Liu}]{luoinstructhoi}
Luo, J.; Ren, W.; Zheng, Q.; Zhang, Y.; Yuan, Z.; Wang, Z.; Lu, H.; and Liu, H. 2025{\natexlab{b}}.
\newblock InstructHOI: Context-Aware Instruction for Multi-Modal Reasoning in Human-Object Interaction Detection.
\newblock In \emph{The Thirty-ninth Annual Conference on Neural Information Processing Systems}.

\bibitem[{Ma et~al.(2021)Ma, Gong, Wang, Huang, Chen, and Yu}]{EPP-MVSNet}
Ma, X.; Gong, Y.; Wang, Q.; Huang, J.; Chen, L.; and Yu, F. 2021.
\newblock Epp-Mvsnet: {{Epipolar-assembling}} Based Depth Prediction for Multi-View Stereo.
\newblock In \emph{Proceedings of the {{IEEE}}/{{CVF International Conference}} on {{Computer Vision}}}, 5732--5740.

\bibitem[{Qian et~al.(2025{\natexlab{a}})Qian, Han, Ding, Lyu, Yuan, Chen, and Liu}]{qian2025adaptive}
Qian, C.; Han, K.; Ding, J.; Lyu, C.; Yuan, Z.; Chen, J.; and Liu, Z. 2025{\natexlab{a}}.
\newblock Adaptive label correction for robust medical image segmentation with noisy labels.
\newblock \emph{arXiv preprint arXiv:2503.12218}.

\bibitem[{Qian et~al.(2025{\natexlab{b}})Qian, Han, Wang, Yuan, Lyu, Chen, and Liu}]{qian2025dyncim}
Qian, C.; Han, K.; Wang, J.; Yuan, Z.; Lyu, C.; Chen, J.; and Liu, Z. 2025{\natexlab{b}}.
\newblock Dyncim: Dynamic curriculum for imbalanced multimodal learning.
\newblock \emph{arXiv preprint arXiv:2503.06456}.

\bibitem[{Qian et~al.(2025{\natexlab{c}})Qian, Xing, Li, Zhao, and Tu}]{qian2025decalign}
Qian, C.; Xing, S.; Li, S.; Zhao, Y.; and Tu, Z. 2025{\natexlab{c}}.
\newblock DecAlign: Hierarchical Cross-Modal Alignment for Decoupled Multimodal Representation Learning.
\newblock \emph{arXiv preprint arXiv:2503.11892}.

\bibitem[{Qian et~al.(2024)Qian, Chen, Lou, Khan, Jin, and Fan}]{qianmaskfactory}
Qian, H.; Chen, Y.; Lou, S.; Khan, F.; Jin, X.; and Fan, D.-P. 2024.
\newblock MaskFactory: Towards High-quality Synthetic Data Generation for Dichotomous Image Segmentation.
\newblock In \emph{NeurIPS}.

\bibitem[{Qu et~al.(2025{\natexlab{a}})Qu, Gou, Zhuang, Yu, Song, Wang, Li, and Xiong}]{ProAPO}
Qu, X.; Gou, G.; Zhuang, J.; Yu, J.; Song, K.; Wang, Q.; Li, Y.; and Xiong, G. 2025{\natexlab{a}}.
\newblock ProAPO: Progressively Automatic Prompt Optimization for Visual Classification.
\newblock In \emph{{IEEE/CVF} Conference on Computer Vision and Pattern Recognition, {CVPR} 2025, Nashville, TN, USA, June 11-15, 2025}, 25145--25155.

\bibitem[{Qu et~al.(2024)Qu, Yu, Gai, Zhuang, Tang, Xiong, Gou, and Wu}]{EmDepart}
Qu, X.; Yu, J.; Gai, K.; Zhuang, J.; Tang, Y.; Xiong, G.; Gou, G.; and Wu, Q. 2024.
\newblock Visual-Semantic Decomposition and Partial Alignment for Document-based Zero-Shot Learning.
\newblock In \emph{Proceedings of the 32nd {ACM} International Conference on Multimedia, {MM} 2024, Melbourne, VIC, Australia, 28 October 2024 - 1 November 2024}, 4581--4590. {ACM}.

\bibitem[{Qu et~al.(2025{\natexlab{b}})Qu, Yu, Zhuang, Gou, Xiong, and Wu}]{MADS}
Qu, X.; Yu, J.; Zhuang, J.; Gou, G.; Xiong, G.; and Wu, Q. 2025{\natexlab{b}}.
\newblock {MADS:} Multi-Attribute Document Supervision for Zero-Shot Image Classification.
\newblock \emph{CoRR}, abs/2503.06847.

\bibitem[{Ren et~al.(2023)Ren, Xu, Zhang, and Yang}]{HPM-MVS}
Ren, C.; Xu, Q.; Zhang, S.; and Yang, J. 2023.
\newblock Hierarchical Prior Mining for Non-Local Multi-View Stereo.
\newblock In \emph{Proceedings of the {{IEEE}}/{{CVF International Conference}} on {{Computer Vision}}}, 3611--3620.

\bibitem[{Sch{\"o}nberger et~al.(2016)Sch{\"o}nberger, Zheng, Frahm, and Pollefeys}]{COLMAP}
Sch{\"o}nberger, J.~L.; Zheng, E.; Frahm, J.-M.; and Pollefeys, M. 2016.
\newblock Pixelwise View Selection for Unstructured Multi-View Stereo.
\newblock In \emph{Proc. Eur. Conf. Comput. Vis. (ECCV)}, 501--518.

\bibitem[{Schops et~al.(2017)Schops, Schonberger, Galliani, Sattler, Schindler, Pollefeys, and Geiger}]{ETH3D}
Schops, T.; Schonberger, J.~L.; Galliani, S.; Sattler, T.; Schindler, K.; Pollefeys, M.; and Geiger, A. 2017.
\newblock A Multi-View Stereo Benchmark With High-Resolution Images and Multi-Camera Videos.
\newblock In \emph{Proc. IEEE/CVF Conf. Comput. Vis. Pattern Recognit. (CVPR)}.

\bibitem[{Shen et~al.(2024{\natexlab{a}})Shen, Jiang, He, Ye, Wang, Du, Li, and Tang}]{Shen2024b}
Shen, F.; Jiang, X.; He, X.; Ye, H.; Wang, C.; Du, X.; Li, Z.; and Tang, J. 2024{\natexlab{a}}.
\newblock {{IMAGDressing-v1}}: {{Customizable Virtual Dressing}}.
\newblock arXiv:2407.12705.

\bibitem[{Shen and Tang(2024)}]{Shen2024}
Shen, F.; and Tang, J. 2024.
\newblock {{IMAGPose}}: {{A Unified Conditional Framework}} for {{Pose-Guided Person Generation}}.
\newblock In \emph{The {{Thirty-eighth Annual Conference}} on {{Neural Information Processing Systems}}}.

\bibitem[{Shen et~al.(2024{\natexlab{b}})Shen, Ye, Liu, Zhang, Wang, Han, and Yang}]{Shen2024c}
Shen, F.; Ye, H.; Liu, S.; Zhang, J.; Wang, C.; Han, X.; and Yang, W. 2024{\natexlab{b}}.
\newblock Boosting {{Consistency}} in {{Story Visualization}} with {{Rich-Contextual Conditional Diffusion Models}}.
\newblock arXiv:2407.02482.

\bibitem[{Shen et~al.(2024{\natexlab{c}})Shen, Ye, Zhang, Wang, Han, and Yang}]{Shen2024a}
Shen, F.; Ye, H.; Zhang, J.; Wang, C.; Han, X.; and Yang, W. 2024{\natexlab{c}}.
\newblock Advancing {{Pose-Guided Image Synthesis}} with {{Progressive Conditional Diffusion Models}}.
\newblock arXiv:2310.06313.

\bibitem[{Shen(2013)}]{Accurate}
Shen, S. 2013.
\newblock Accurate Multiple View 3D Reconstruction Using Patch-Based Stereo for Large-Scale Scenes.
\newblock \emph{IEEE Trans. Image Process.}, 22(5): 1901--1914.

\bibitem[{Shen et~al.(2025)Shen, Liu, Feng, Ma, and An}]{shen2025topology}
Shen, T.; Liu, S.; Feng, J.; Ma, Z.; and An, N. 2025.
\newblock Topology-Aware 3D Gaussian Splatting: Leveraging Persistent Homology for Optimized Structural Integrity.
\newblock In \emph{Proceedings of the AAAI Conference on Artificial Intelligence}, volume~39, 6823--6832.

\bibitem[{Song et~al.(2025{\natexlab{a}})Song, Li, Dun, Huang, Cao, and Ye}]{song2025gpromptshield}
Song, S.; Li, P.; Dun, M.; Huang, M.; Cao, H.; and Ye, X. 2025{\natexlab{a}}.
\newblock GPromptShield: Elevating Resilience in Graph Prompt Tuning Against Adversarial Attacks.
\newblock In \emph{The Thirteenth International Conference on Learning Representations}.

\bibitem[{Song et~al.(2025{\natexlab{b}})Song, Li, Dun, Zhang, Cao, and Ye}]{song2025equipping}
Song, S.; Li, P.; Dun, M.; Zhang, Y.; Cao, H.; and Ye, X. 2025{\natexlab{b}}.
\newblock Equipping Graph Autoencoders: Revisiting Masking Strategies from a Robustness Perspective.
\newblock In \emph{Proceedings of the 2025 SIAM International Conference on Data Mining (SDM)}, 366--375. SIAM.

\bibitem[{Song et~al.(2025{\natexlab{c}})Song, Li, Dun, Zhang, Cao, and Ye}]{song2025spmgae}
Song, S.; Li, P.; Dun, M.; Zhang, Y.; Cao, H.; and Ye, X. 2025{\natexlab{c}}.
\newblock SPMGAE: Self-purified masked graph autoencoders release robust expression power.
\newblock \emph{Neurocomputing}, 611: 128631.

\bibitem[{Song et~al.(2023)Song, Bi, Zhang, Mao, and Wang}]{song2023xvtp3d}
Song, Z.; Bi, H.; Zhang, R.; Mao, T.; and Wang, Z. 2023.
\newblock Xvtp3d: cross-view trajectory prediction using shared 3d queries for autonomous driving.
\newblock \emph{arXiv preprint arXiv:2308.08764}.

\bibitem[{Song et~al.(2025{\natexlab{d}})Song, Lin, Huang, Wang, and Lin}]{song2025siri}
Song, Z.; Lin, X.; Huang, Q.; Wang, G.; and Lin, L. 2025{\natexlab{d}}.
\newblock SIRI-Bench: Challenging VLMs' Spatial Intelligence through Complex Reasoning Tasks.
\newblock \emph{arXiv preprint arXiv:2506.14512}.

\bibitem[{Song et~al.(2025{\natexlab{e}})Song, Qin, Chen, Lin, and Wang}]{song2025physical}
Song, Z.; Qin, S.; Chen, T.; Lin, L.; and Wang, G. 2025{\natexlab{e}}.
\newblock Physical autoregressive model for robotic manipulation without action pretraining.
\newblock \emph{arXiv preprint arXiv:2508.09822}.

\bibitem[{Sun et~al.(2024)Sun, Xu, Jin, Luo, Qian, and Liu}]{SUN2}
Sun, H.; Xu, L.; Jin, S.; Luo, P.; Qian, C.; and Liu, W. 2024.
\newblock PROGRAM: PROtotype GRAph Model based Pseudo-Label Learning for Test-Time Adaptation.
\newblock In \emph{The Twelfth International Conference on Learning Representations}.

\bibitem[{Sun et~al.(2025{\natexlab{a}})Sun, Fang, Lu, Zhao, and Ling}]{sun2025end}
Sun, N.; Fang, H.; Lu, Y.; Zhao, C.; and Ling, H. 2025{\natexlab{a}}.
\newblock END2: Robust Dual-Decoder Watermarking Framework Against Non-Differentiable Distortions.
\newblock In \emph{Proceedings of the AAAI Conference on Artificial Intelligence}, volume~39, 773--781.

\bibitem[{Sun et~al.(2025{\natexlab{b}})Sun, Yuan, Fang, Lu, Ling, Xie, and Zhao}]{sunultra}
Sun, N.; Yuan, L.; Fang, H.; Lu, Y.; Ling, H.; Xie, S.; and Zhao, C. 2025{\natexlab{b}}.
\newblock Ultra-high Resolution Watermarking Framework Resistant to Extreme Cropping and Scaling.
\newblock In \emph{The Thirty-ninth Annual Conference on Neural Information Processing Systems}.

\bibitem[{Sun et~al.(2022)Sun, Liu, Li, Ying, Zhai, and Mou}]{API-MVS}
Sun, S.; Liu, J.; Li, Y.; Ying, H.; Zhai, Z.; and Mou, Y. 2022.
\newblock Adaptive Pixelwise Inference Multi-View Stereo.
\newblock In Xu, D.; and Xiao, L., eds., \emph{Thirteenth {{International Conference}} on {{Graphics}} and {{Image Processing}} ({{ICGIP}} 2021)}, 77. Kunming, China: SPIE.
\newblock ISBN 978-1-5106-5042-8 978-1-5106-5043-5.

\bibitem[{Sun et~al.(2021)Sun, Zheng, Shi, Xu, and Liu}]{PHI-MVS}
Sun, S.; Zheng, Y.; Shi, X.; Xu, Z.; and Liu, Y. 2021.
\newblock Phi-Mvs: {{Plane}} Hypothesis Inference Multi-View Stereo for Large-Scale Scene Reconstruction.
\newblock \emph{arXiv preprint arXiv:2104.06165}.

\bibitem[{Tang et~al.(2024)Tang, Cao, Hou, Jiang, Liu, and Meng}]{tang2024crs}
Tang, D.; Cao, X.; Hou, X.; Jiang, Z.; Liu, J.; and Meng, D. 2024.
\newblock Crs-diff: Controllable remote sensing image generation with diffusion model.
\newblock \emph{IEEE Transactions on Geoscience and Remote Sensing}.

\bibitem[{Tang et~al.(2025)Tang, Cao, Wu, Li, Yao, Bai, Jiang, Li, and Meng}]{tang2025aerogen}
Tang, D.; Cao, X.; Wu, X.; Li, J.; Yao, J.; Bai, X.; Jiang, D.; Li, Y.; and Meng, D. 2025.
\newblock AeroGen: Enhancing remote sensing object detection with diffusion-driven data generation.
\newblock In \emph{Proceedings of the Computer Vision and Pattern Recognition Conference}, 3614--3624.

\bibitem[{Wang et~al.(2022{\natexlab{a}})Wang, Galliani, Vogel, and Pollefeys}]{IterMVS}
Wang, F.; Galliani, S.; Vogel, C.; and Pollefeys, M. 2022{\natexlab{a}}.
\newblock IterMVS: Iterative probability estimation for efficient multi-view stereo.
\newblock In \emph{Proc. IEEE/CVF Conf. Comput. Vis. Pattern Recognit. (CVPR)}, 8606--8615.

\bibitem[{Wang et~al.(2025{\natexlab{a}})Wang, Chen, Liu, Liu, Liu, Gao, and Xiong}]{wang2025masktwins}
Wang, J.; Chen, Y.; Liu, X.; Liu, C.; Liu, D.; Gao, J.; and Xiong, Z. 2025{\natexlab{a}}.
\newblock MaskTwins: Dual-form Complementary Masking for Domain-Adaptive Image Segmentation.
\newblock In \emph{International Conference on Machine Learning (ICML)}.

\bibitem[{Wang et~al.(2025{\natexlab{b}})Wang, Lin, Guan, Nie, He, Li, Liao, and Zhao}]{wang2025jasmine}
Wang, J.; Lin, C.; Guan, C.; Nie, L.; He, J.; Li, H.; Liao, K.; and Zhao, Y. 2025{\natexlab{b}}.
\newblock Jasmine: Harnessing Diffusion Prior for Self-supervised Depth Estimation.
\newblock \emph{arXiv preprint arXiv:2503.15905}.

\bibitem[{Wang et~al.(2024{\natexlab{a}})Wang, Lin, Nie, Huang, Zhao, Pan, and Ai}]{Wang_2024}
Wang, J.; Lin, C.; Nie, L.; Huang, S.; Zhao, Y.; Pan, X.; and Ai, R. 2024{\natexlab{a}}.
\newblock WeatherDepth: Curriculum Contrastive Learning for Self-Supervised Depth Estimation under Adverse Weather Conditions.
\newblock In \emph{2024 IEEE International Conference on Robotics and Automation (ICRA)}, 4976–4982. IEEE.

\bibitem[{Wang et~al.(2024{\natexlab{b}})Wang, Lin, Nie, Liao, Shao, and Zhao}]{Wang_2024_2}
Wang, J.; Lin, C.; Nie, L.; Liao, K.; Shao, S.; and Zhao, Y. 2024{\natexlab{b}}.
\newblock Digging into Contrastive Learning for Robust Depth Estimation with Diffusion Models.
\newblock In \emph{Proceedings of the 32nd ACM International Conference on Multimedia}, 4129–4137. ACM.

\bibitem[{Wang et~al.(2025{\natexlab{c}})Wang, Lin, Sun, Liu, Nie, Li, Liao, Chu, and Zhao}]{wang2025editor}
Wang, J.; Lin, C.; Sun, L.; Liu, R.; Nie, L.; Li, M.; Liao, K.; Chu, X.; and Zhao, Y. 2025{\natexlab{c}}.
\newblock From Editor to Dense Geometry Estimator.
\newblock \emph{arXiv preprint arXiv:2509.04338}.

\bibitem[{Wang et~al.(2025{\natexlab{d}})Wang, Chen, Xiao, Xiao, Li, Chen, Ye, Xu, Zhang, Yan et~al.}]{wang2025unifying}
Wang, N.; Chen, Y.; Xiao, L.; Xiao, W.; Li, B.; Chen, Z.; Ye, C.; Xu, S.; Zhang, S.; Yan, Z.; et~al. 2025{\natexlab{d}}.
\newblock Unifying Appearance Codes and Bilateral Grids for Driving Scene Gaussian Splatting.
\newblock \emph{arXiv preprint arXiv:2506.05280}.

\bibitem[{Wang et~al.(2022{\natexlab{b}})Wang, Zhu, Huang, Qin, Ye, He, Chi, and Wang}]{MVSTER}
Wang, X.; Zhu, Z.; Huang, G.; Qin, F.; Ye, Y.; He, Y.; Chi, X.; and Wang, X. 2022{\natexlab{b}}.
\newblock {{MVSTER}}: {{Epipolar}} Transformer for Efficient Multi-View Stereo.
\newblock In \emph{European {{Conference}} on {{Computer Vision}}}, 573--591. Springer.

\bibitem[{Wang et~al.(2020)Wang, Guan, Chen, Luo, Luo, and Ju}]{MG-MVS}
Wang, Y.; Guan, T.; Chen, Z.; Luo, Y.; Luo, K.; and Ju, L. 2020.
\newblock Mesh-Guided Multi-View Stereo With Pyramid Architecture.
\newblock In \emph{Proc. IEEE/CVF Conf. Comput. Vis. Pattern Recognit. (CVPR)}, 2036--2045.

\bibitem[{Wang et~al.(2023)Wang, Zeng, Guan, Yang, Chen, Liu, Xu, and Luo}]{APD-MVS}
Wang, Y.; Zeng, Z.; Guan, T.; Yang, W.; Chen, Z.; Liu, W.; Xu, L.; and Luo, Y. 2023.
\newblock Adaptive Patch Deformation for Textureless-Resilient Multi-View Stereo.
\newblock In \emph{Proc. IEEE/CVF Conf. Comput. Vis. Pattern Recognit. (CVPR)}, 1621--1630.

\bibitem[{Xing et~al.(2025)Xing, Li, Wang, Bai, Wang, Hu, Qian, Yao, and Tu}]{xing2025re}
Xing, S.; Li, P.; Wang, Y.; Bai, R.; Wang, Y.; Hu, C.-W.; Qian, C.; Yao, H.; and Tu, Z. 2025.
\newblock Re-Align: Aligning vision language models via retrieval-augmented direct preference optimization.
\newblock In \emph{Proceedings of the 2025 Conference on Empirical Methods in Natural Language Processing}, 2379--2397.

\bibitem[{Xu et~al.(2022)Xu, Kong, Tao, and Pollefeys}]{ACMMP}
Xu, Q.; Kong, W.; Tao, W.; and Pollefeys, M. 2022.
\newblock Multi-{{Scale Geometric Consistency Guided}} and {{Planar Prior Assisted Multi-View Stereo}}.
\newblock \emph{IEEE Transactions on Pattern Analysis and Machine Intelligence}.

\bibitem[{Xu and Tao(2019)}]{ACMM}
Xu, Q.; and Tao, W. 2019.
\newblock Multi-Scale Geometric Consistency Guided Multi-View Stereo.
\newblock In \emph{Proc. IEEE/CVF Conf. Comput. Vis. Pattern Recognit. (CVPR)}.

\bibitem[{Xu et~al.(2020)Xu, Liu, Shi, Wang, and Zheng}]{MAR-MVS}
Xu, Z.; Liu, Y.; Shi, X.; Wang, Y.; and Zheng, Y. 2020.
\newblock {MARMVS:} Matching Ambiguity Reduced Multiple View Stereo for Efficient Large Scale Scene Reconstruction.
\newblock In \emph{Proc. IEEE/CVF Conf. Comput. Vis. Pattern Recognit. (CVPR)}, 5980--5989.

\bibitem[{Yan et~al.(2024{\natexlab{a}})Yan, Lv, Yang, Lin, Zheng, and Zhang}]{10226279}
Yan, R.; Lv, Z.; Yang, Z.; Lin, S.; Zheng, C.; and Zhang, F. 2024{\natexlab{a}}.
\newblock Sparse and Hierarchical Transformer for Survival Analysis on Whole Slide Images.
\newblock \emph{IEEE Journal of Biomedical and Health Informatics}, 28(1): 7--18.

\bibitem[{Yan et~al.(2025)Yan, Dong, Shao, Lu, Liu, Liu, Wang, Wang, Wang, Remondino et~al.}]{yan2025renderworld}
Yan, Z.; Dong, W.; Shao, Y.; Lu, Y.; Liu, H.; Liu, J.; Wang, H.; Wang, Z.; Wang, Y.; Remondino, F.; et~al. 2025.
\newblock Renderworld: World model with self-supervised 3d label.
\newblock In \emph{2025 IEEE International Conference on Robotics and Automation (ICRA)}, 6063--6070. IEEE.

\bibitem[{Yan et~al.(2024{\natexlab{b}})Yan, Li, Shao, Chen, Wu, Hwang, Zhao, and Remondino}]{yan20243dsceneeditor}
Yan, Z.; Li, L.; Shao, Y.; Chen, S.; Wu, Z.; Hwang, J.-N.; Zhao, H.; and Remondino, F. 2024{\natexlab{b}}.
\newblock 3dsceneeditor: Controllable 3d scene editing with gaussian splatting.
\newblock \emph{arXiv preprint arXiv:2412.01583}.

\bibitem[{Yan et~al.(2023)Yan, Mazzacca, Rigon, Farella, Trybala, Remondino et~al.}]{yan2023nerfbk}
Yan, Z.; Mazzacca, G.; Rigon, S.; Farella, E.~M.; Trybala, P.; Remondino, F.; et~al. 2023.
\newblock NeRFBK: a holistic dataset for benchmarking NeRF-based 3D reconstruction.
\newblock \emph{International Archives of the Photogrammetry, Remote Sensing and Spatial Information Sciences}, 48(1): 219--226.

\bibitem[{Yang et~al.(2020)Yang, Mao, Alvarez, and Liu}]{Cas-MVSNet}
Yang, J.; Mao, W.; Alvarez, J.~M.; and Liu, M. 2020.
\newblock Cost Volume Pyramid Based Depth Inference for Multi-View Stereo.
\newblock In \emph{Proc. IEEE/CVF Conf. Comput. Vis. Pattern Recognit. (CVPR)}, 4876--4885.

\bibitem[{Yang et~al.(2024{\natexlab{a}})Yang, Kang, Huang, Zhao, Xu, Feng, and Zhao}]{depany2}
Yang, L.; Kang, B.; Huang, Z.; Zhao, Z.; Xu, X.; Feng, J.; and Zhao, H. 2024{\natexlab{a}}.
\newblock Depth {{Anything V2}}.
\newblock arXiv:2406.09414.

\bibitem[{Yang et~al.(2024{\natexlab{b}})Yang, Wang, Zhang, Zhang, Zhi, Zhao, Kong, Zhou, and Zhou}]{10858860}
Yang, Y.; Wang, Z.; Zhang, D.; Zhang, H.; Zhi, P.; Zhao, R.; Kong, X.; Zhou, R.; and Zhou, Q. 2024{\natexlab{b}}.
\newblock A3Framework: A Cloud Framework with Autonomous Driving Path Planning.
\newblock In \emph{2024 Twelfth International Conference on Advanced Cloud and Big Data (CBD)}, 136--141.

\bibitem[{Yang et~al.(2024{\natexlab{c}})Yang, Li, Zang, Han, and Zhang}]{cmb20240513}
Yang, Z.; Li, H.; Zang, D.; Han, R.; and Zhang, F. 2024{\natexlab{c}}.
\newblock Improved Denoising of Cryo-Electron Microscopy Micrographs with Simulation-Aware Pretraining.
\newblock \emph{Journal of Computational Biology}, 31(6): 564--575.

\bibitem[{Yang et~al.(2025)Yang, Shi, Ba, Song, Luan, Hu, Lin, Wang, Zhou, and Yan}]{yang2025fusionmultiscaleheterogeneouspathology}
Yang, Z.; Shi, X.; Ba, W.; Song, Z.; Luan, H.; Hu, T.; Lin, S.; Wang, J.; Zhou, S.~K.; and Yan, R. 2025.
\newblock Fusion of Multi-scale Heterogeneous Pathology Foundation Models for Whole Slide Image Analysis.

\bibitem[{Yang et~al.(2024{\natexlab{d}})Yang, Zang, Li, Zhang, Zhang, and Han}]{Yang2024c}
Yang, Z.; Zang, D.; Li, H.; Zhang, Z.; Zhang, F.; and Han, R. 2024{\natexlab{d}}.
\newblock Self-Supervised Noise Modeling and Sparsity Guided Electron Tomography Volumetric Image Denoising.
\newblock \emph{Ultramicroscopy}, 255: 113860.

\bibitem[{Yang et~al.(2024{\natexlab{e}})Yang, Zang, Li, Zhang, Zhang, and Han}]{nmsg}
Yang, Z.; Zang, D.; Li, H.; Zhang, Z.; Zhang, F.; and Han, R. 2024{\natexlab{e}}.
\newblock Self-supervised noise modeling and sparsity guided electron tomography volumetric image denoising.
\newblock \emph{Ultramicroscopy}, 255: 113860.

\bibitem[{Yang, Zhang, and Han(2021{\natexlab{a}})}]{Yang2021}
Yang, Z.; Zhang, F.; and Han, R. 2021{\natexlab{a}}.
\newblock Self-Supervised Cryo-Electron Tomography Volumetric Image Restoration from Single Noisy Volume with Sparsity Constraint.
\newblock In \emph{Proceedings of the {{IEEE}}/{{CVF International Conference}} on {{Computer Vision}}}, 4056--4065.

\bibitem[{Yang, Zhang, and Han(2021{\natexlab{b}})}]{Yang_2021_ICCV}
Yang, Z.; Zhang, F.; and Han, R. 2021{\natexlab{b}}.
\newblock Self-Supervised Cryo-Electron Tomography Volumetric Image Restoration From Single Noisy Volume With Sparsity Constraint.
\newblock In \emph{Proceedings of the IEEE/CVF International Conference on Computer Vision (ICCV)}, 4056--4065.

\bibitem[{Yao et~al.(2018)Yao, Luo, Li, Fang, and Quan}]{MVSNet}
Yao, Y.; Luo, Z.; Li, S.; Fang, T.; and Quan, L. 2018.
\newblock MVSNet: Depth Inference for Unstructured Multi-view Stereo.
\newblock In \emph{Proc. Eur. Conf. Comput. Vis. (ECCV)}.

\bibitem[{Yao et~al.(2020)Yao, Luo, Li, Zhang, Ren, Zhou, Fang, and Quan}]{Blendedmvs}
Yao, Y.; Luo, Z.; Li, S.; Zhang, J.; Ren, Y.; Zhou, L.; Fang, T.; and Quan, L. 2020.
\newblock Blendedmvs: {{A}} Large-Scale Dataset for Generalized Multi-View Stereo Networks.
\newblock In \emph{Proceedings of the {{IEEE}}/{{CVF Conference}} on {{Computer Vision}} and {{Pattern Recognition}}}, 1790--1799.

\bibitem[{Yuan et~al.(2024{\natexlab{a}})Yuan, Cao, Li, Jiang, and Wang}]{SD-MVS}
Yuan, Z.; Cao, J.; Li, Z.; Jiang, H.; and Wang, Z. 2024{\natexlab{a}}.
\newblock {{SD-MVS}}: {{Segmentation-driven}} Deformation Multi-View Stereo with Spherical Refinement and {{EM}} Optimization.
\newblock \emph{Proceedings of the AAAI Conference on Artificial Intelligence}, 38(7): 6871--6880.

\bibitem[{Yuan et~al.(2024{\natexlab{b}})Yuan, Cao, Wang, and Li}]{TSAR-MVS}
Yuan, Z.; Cao, J.; Wang, Z.; and Li, Z. 2024{\natexlab{b}}.
\newblock Tsar-Mvs: {{Textureless-aware}} Segmentation and Correlative Refinement Guided Multi-View Stereo.
\newblock \emph{Pattern Recognition}, 110565.

\bibitem[{Yuan et~al.(2025{\natexlab{a}})Yuan, Qu, Qian, Chen, Tang, Sun, Chu, Zhang, Wang, Cai et~al.}]{yuan2025video}
Yuan, Z.; Qu, X.; Qian, C.; Chen, R.; Tang, J.; Sun, L.; Chu, X.; Zhang, D.; Wang, Y.; Cai, Y.; et~al. 2025{\natexlab{a}}.
\newblock Video-star: Reinforcing open-vocabulary action recognition with tools.
\newblock \emph{arXiv preprint arXiv:2510.08480}.

\bibitem[{Yuan et~al.(2025{\natexlab{b}})Yuan, Tang, Luo, Chen, Qian, Sun, Chu, Cai, Zhang, and Li}]{yuan2025autodrive}
Yuan, Z.; Tang, J.; Luo, J.; Chen, R.; Qian, C.; Sun, L.; Chu, X.; Cai, Y.; Zhang, D.; and Li, S. 2025{\natexlab{b}}.
\newblock AutoDrive-R2: Incentivizing Reasoning and Self-Reflection Capacity for VLA Model in Autonomous Driving.
\newblock \emph{arXiv preprint arXiv:2509.01944}.

\bibitem[{Yuan et~al.(2025{\natexlab{c}})Yuan, Yang, Cai, Wu, Liu, Zhang, Jiang, Li, and Wang}]{yuan2025sed}
Yuan, Z.; Yang, Z.; Cai, Y.; Wu, K.; Liu, M.; Zhang, D.; Jiang, H.; Li, Z.; and Wang, Z. 2025{\natexlab{c}}.
\newblock SED-MVS: Segmentation-Driven and Edge-Aligned Deformation Multi-View Stereo with Depth Restoration and Occlusion Constraint.
\newblock \emph{IEEE Transactions on Circuits and Systems for Video Technology}.

\bibitem[{Zhai et~al.(2024{\natexlab{a}})Zhai, {\"O}rnek, Chen, Liao, Di, Navab, Tombari, and Busam}]{YZF_2}
Zhai, G.; {\"O}rnek, E.~P.; Chen, D.~Z.; Liao, R.; Di, Y.; Navab, N.; Tombari, F.; and Busam, B. 2024{\natexlab{a}}.
\newblock EchoScene: Indoor Scene Generation via Information Echo over Scene Graph Diffusion.
\newblock \emph{arXiv preprint arXiv:2405.00915}.

\bibitem[{Zhai et~al.(2024{\natexlab{b}})Zhai, {\"O}rnek, Wu, Di, Tombari, Navab, and Busam}]{YZF_1}
Zhai, G.; {\"O}rnek, E.~P.; Wu, S.-C.; Di, Y.; Tombari, F.; Navab, N.; and Busam, B. 2024{\natexlab{b}}.
\newblock Commonscenes: Generating commonsense 3d indoor scenes with scene graphs.
\newblock \emph{Advances in Neural Information Processing Systems}, 36.

\bibitem[{Zhang et~al.(2024{\natexlab{a}})Zhang, Cheng, Zhang, and Liu}]{Zhang2}
Zhang, B.; Cheng, D.; Zhang, Y.; and Liu, F. 2024{\natexlab{a}}.
\newblock {{FP}}={{xINT}}:{{A Low-Bit Series Expansion Algorithm}} for {{Post-Training Quantization}}.
\newblock arXiv:2412.06865.

\bibitem[{Zhang et~al.(2024{\natexlab{b}})Zhang, Cheng, Zhang, Liu, and Chen}]{Zhang1}
Zhang, B.; Cheng, D.; Zhang, Y.; Liu, F.; and Chen, W. 2024{\natexlab{b}}.
\newblock Compression for {{Better}}: {{A General}} and {{Stable Lossless Compression Framework}}.
\newblock arXiv:2412.06868.

\bibitem[{Zhang et~al.(2024{\natexlab{c}})Zhang, Cheng, Zhang, Liu, and Tian}]{Zhang3}
Zhang, B.; Cheng, D.; Zhang, Y.; Liu, F.; and Tian, J. 2024{\natexlab{c}}.
\newblock Lossless {{Model Compression}} via {{Joint Low-Rank Factorization Optimization}}.
\newblock arXiv:2412.06867.

\bibitem[{Zhang et~al.(2025{\natexlab{a}})Zhang, Chen, Zhi, Chen, Yuan, Li, Zhou, Zhou et~al.}]{zhang2025mapexpert}
Zhang, D.; Chen, D.; Zhi, P.; Chen, Y.; Yuan, Z.; Li, C.; Zhou, R.; Zhou, Q.; et~al. 2025{\natexlab{a}}.
\newblock Mapexpert: Online hd map construction with simple and efficient sparse map element expert.
\newblock In \emph{Proceedings of the AAAI Conference on Artificial Intelligence}, volume~39, 14745--14753.

\bibitem[{Zhang et~al.(2025{\natexlab{b}})Zhang, Sun, Hu, Wu, Yuan, Zhou, Shen, and Zhou}]{zhang2025pure}
Zhang, D.; Sun, J.; Hu, C.; Wu, X.; Yuan, Z.; Zhou, R.; Shen, F.; and Zhou, Q. 2025{\natexlab{b}}.
\newblock Pure Vision Language Action (VLA) Models: A Comprehensive Survey.
\newblock \emph{arXiv preprint arXiv:2509.19012}.

\bibitem[{Zhang et~al.(2025{\natexlab{c}})Zhang, Yuan, Huang, Yan, Li, Nie, Zhao, Zhou, and Zhou}]{zhangdrive}
Zhang, D.; Yuan, Z.; Huang, K.; Yan, Y.; Li, C.; Nie, H.; Zhao, S.; Zhou, R.; and Zhou, Q. 2025{\natexlab{c}}.
\newblock AT-Drive: Exploiting Adversarial Transfer for End-to-end Autonomous Driving.

\bibitem[{Zhang et~al.(2025{\natexlab{d}})Zhang, Yuan, Li, Chen, Zhao, Nie, Zhou, and Zhou}]{zhangaddi}
Zhang, D.; Yuan, Z.; Li, C.; Chen, Y.; Zhao, S.; Nie, H.; Zhou, R.; and Zhou, Q. 2025{\natexlab{d}}.
\newblock ADDI: A Simplified E2E Autonomous Driving Model with Distinct Experts and Implicit Interactions.

\bibitem[{Zhang et~al.(2023{\natexlab{a}})Zhang, Zhi, Yong, Wang, Hou, Guo, Zhou, and Zhou}]{Zhang2023EHSSAE}
Zhang, D.; Zhi, P.; Yong, B.; Wang, J.-Q.; Hou, Y.; Guo, L.; Zhou, Q.; and Zhou, R. 2023{\natexlab{a}}.
\newblock EHSS: An Efficient Hybrid-supervised Symmetric Stereo Matching Network.
\newblock \emph{2023 IEEE 26th International Conference on Intelligent Transportation Systems (ITSC)}, 1044--1051.

\bibitem[{Zhang, Zhu, and Lin(2023)}]{RA-MVSNet}
Zhang, Y.; Zhu, J.; and Lin, L. 2023.
\newblock Multi-{{View Stereo Representation Revist}}: {{Region-Aware MVSNet}}.
\newblock In \emph{Proceedings of the {{IEEE}}/{{CVF Conference}} on {{Computer Vision}} and {{Pattern Recognition}}}, 17376--17385.

\bibitem[{Zhang et~al.(2023{\natexlab{b}})Zhang, Peng, Hu, and Wang}]{GeoMVSNet}
Zhang, Z.; Peng, R.; Hu, Y.; and Wang, R. 2023{\natexlab{b}}.
\newblock {{GeoMVSNet}}: {{Learning Multi-View Stereo With Geometry Perception}}.
\newblock In \emph{Proceedings of the {{IEEE}}/{{CVF Conference}} on {{Computer Vision}} and {{Pattern Recognition}}}, 21508--21518.

\end{thebibliography}

\end{document}